%% file: main.tex
\documentclass[10pt]{IEEEtran}

\input{notations}

\title{Formal description of ML models for unambiguous implementation}           
\author{
  \IEEEauthorblockN{Adrien Gauffriau}
  \IEEEauthorblockA{\textit{Airbus, France}}
  \and
   \IEEEauthorblockN{Iryna De Albuquerque Silva}
  \IEEEauthorblockA{\textit{ONERA, France}}
  \and
  \IEEEauthorblockN{Claire Pagetti}
  \IEEEauthorblockA{\textit{ONERA, France}}
}

\begin{document}
\maketitle

\input{introduction}

\input{nnef}
\input{distribution}

\input{extsem.tex}

\input{implem.tex}
\input{related-work}

\input{conclusion}

{\footnotesize
\subsection*{\textbf{Acknowledgments}}
This work has benefited from the AI Interdisciplinary Institute ANITI,
funded by the  “Investing for the Future – PIA3” program
Grant agreement ANR-19-P3IA-0004 and
from  the PHYLOG 2 project funded by the French government through the France Relance program, based on the funding from and by the European Union through the NextGenerationEU program.}

\bibliographystyle{alpha}
\bibliography{bib}

\end{document}

%% file: notations.tex
\usepackage[utf8]{inputenc}  
\usepackage[english]{babel}                             
\usepackage{syntax}
\usepackage{tikz}
\usepackage{soul,hyperref}
\usepackage{xcolor,xspace}
\usepackage[framemethod=TikZ]{mdframed}
\usetikzlibrary{petri,positioning,calc,shapes,fit}
\usetikzlibrary {automata} 
\usepackage{listings}
\usepackage[inline]{enumitem}
\usepackage{amsfonts}
\usepackage{amsmath}

\newtheorem{definition}{\textbf{Definition}}
\newtheorem{property}{\textbf{Property}}
\newtheorem{example}{\textbf{Example}}
\newtheorem{remark}{\textbf{Remark}}
\newtheorem{semantic}{\textbf{Semantics}}
\newtheorem{translation}{\textbf{Translation}}

\mdfdefinestyle{MyFrame}{%
     outerlinewidth=-1pt,
    innertopmargin=5pt,
    innerbottommargin=5pt,
    innerrightmargin=10pt,
    innerleftmargin=10pt,
        leftmargin = -1pt,
        rightmargin = -1pt
        }

\definecolor{navyblue}{rgb}{0.0, 0.0, 0.5}

\newcounter{stxcnt}
\newenvironment{mysyntax}[1][]{%
    \refstepcounter{stxcnt}%
    \mdfsetup{%
    frametitle={%
        \tikz[baseline=(current bounding box.east),outer sep=0pt]
        \node[anchor=east,rectangle,rounded corners,fill=navyblue, text=white]
        {\normalsize Syntax\thestxcnt \hspace{1cm}#1};},
    innertopmargin=-1pt,linecolor=navyblue,backgroundcolor=navyblue!10,%
    linewidth=2pt,topline=true,roundcorner=10pt,%
    frametitleaboveskip=\dimexpr-\ht\strutbox\relax%
    }
    \vspace{0.08cm}
    \begin{mdframed}[style=MyFrame,nobreak=true]\relax\footnotesize}{
    \end{mdframed}
    \vspace{-0.3cm}
}

\newcommand{\Fn}{\ensuremath{F_\mathcal{N}}\xspace}
\newcommand{\petri}{Petri\xspace}
\newcommand{\scade}{{\sc scade}\xspace}
\newcommand{\nnef}{{\sc nnef}\xspace}
\newcommand{\xtratum}{{\sc xtratum}\xspace}
\newcommand{\reluplex}{{\sc reluplex}\xspace}
\newcommand{\onnx}{{\sc onnx}\xspace}
\newcommand{\cuda}{{\sc cuda}\xspace}
\newcommand{\cudnn}{{\sc cudnn}\xspace}
\newcommand{\lenet}{{\sc LeNet-5}\xspace}
\newcommand{\khronos}{{\sc Khronos}\xspace}
\newcommand{\arm}{{\sc arm}\xspace}
\newcommand{\gpu}{{\sc gpu}\xspace}
\newcommand{\gpus}{{\sc gpu}s\xspace}
\newcommand{\fpga}{{\sc fpga}\xspace}
\newcommand{\nvidia}{{\sc nvidia}\xspace}
\newcommand{\ultrascale}{{\sc UltraScale+}\xspace}
\newcommand{\xavier}{{\sc Xavier}\xspace}
\newcommand{\pytorch}{{\sc PyTorch}\xspace}
\newcommand{\python}{{\sc Python}\xspace}
\newcommand{\tensorflow}{{\sc TensorFlow}\xspace}

\newcommand{\posix}{{\sc posix}\xspace}
\newcommand{\keras}{{\sc Keras}\xspace}

\usetikzlibrary{arrows.meta}

\newcommand*{\medcap}{\mathbin{\scalebox{1.5}{\ensuremath{\cap}}}}
\newcommand*{\medcup}{\mathbin{\scalebox{1.5}{\ensuremath{\cup}}}}

%% file: introduction.tex
\section{Introduction}
Machine learning (ML) applications have been gaining considerable attention in the field of transportation. However, their use in real-life operational safety-critical products, in particular in the aeronautical domain subject to stringent certification, raises several issues regarding functional correctness, compliance with requirements, formal verification, safety or implementation.
In order to tackle those issues,
new guidelines -- named  ED 324/ ARP 6983 standard \cite{wg114} --
are currently drafted by the EUROCAE WG-114/SAE G-34 joint working group
that cover the whole spectrum of the system development including the data sets composition, the ML model design and its implementation. In this paper, we focus only on the implementation of the ML model.

\subsection{Context}
\begin{figure}[hbt]
 \centering
  \resizebox{.95\linewidth}{!}{\input{currentpractice}}
  \caption{Left part: current ML deployment practice and right part: proposed aeronautical practice.\label{fig:currentpractice}}
\end{figure}
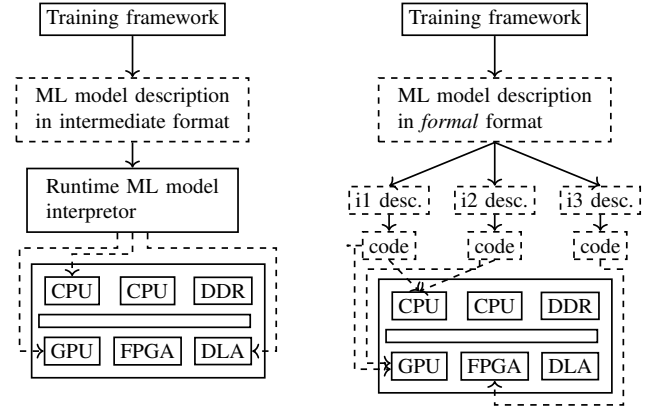

In the ML current practices, a \emph{training framework} is used to design an ML model and the resulting ML model
is then deployed on the target with a \emph{deployment framework}.
It is up to the training framework or a designer to export the trained model description
in an exchange format and up to the deployment
framework to import the ML model description.
The left part of figure \ref{fig:currentpractice} shows those practices.
The deployment framework is most of the time an ML model interpreter,
that can accommodate any type of ML model architecture,
and that allocates at runtime the execution on the different
available accelerators (e.g. \gpu, \fpga) of the target.
These ML frameworks have been designed 1) to ease as
much as possible the deployment of models for the users
and 2) to optimize
as much as possible the execution performance (usually expressed
in trillion operations per second -- TeraOp/s).
As a result, they are very impressive and
allow for complex deployments and optimizations. Those are
hard, if not impossible, to reproduce for a programmer without
using ML libraries (e.g. \cudnn on \nvidia).

The counterpart of such an approach is
1) the absence or limited information of
internal computation and allocation;
2) small modifications and adaptations
of the ML model
(e.g. when exporting the ML model description
or
when quantizing on the fly some
matrix multiplication to execute on deep learning accelerator -- DLA).
If this grey/black box approach
is acceptable and suitable for mainstream applications,
it is a blocking point for highly safety-critical applications.
As a result, the main objective
of the ARP 6983 standard with respect to the implementation process
is the \emph{semantics preservation} of the off-line trained model on the final hardware platform.
This means that the execution of the ML model in the training
framework should be exactly reproduced on the embedded target during execution.
To reach that objective, an alternative development process is proposed
as illustrated in the right hand part of figure \ref{fig:currentpractice}.


\subsection{Assurance development process}
The principle of this trustable development process (figure \ref{fig:currentpractice}) is to ensure the semantic preservation at each step of the development cycle (e.g. between training framework and description, between description and code, between code and executable).
This is guaranteed thanks to a series of requirements provided in the ARP 6983.

\textbf{Requirement 1.}
First, the trained model must be formally
and unambiguously defined in what we call an adequate \emph{format}.
Such a \emph{format}
must come with a formal syntax and semantic,
and should be agnostic of any (training and/or deployment) framework.

\textbf{Requirement 2.}
Second, the implementation process
should allow several deployments on hardware platforms
and it must be known beforehand how the ML model will be mapped
on the accelerators and when.
This entails in particular that
the  \emph{format} should allow several types of deployment such as
distribution, parallelization or pipelining.
Thus, there is a step that consists in splitting the
ML model description (in the chosen format) as a series of \emph{item} descriptions.
Indeed, in the avionics domain,
a target processor is decomposed as a set of software (SW) and/or hardware (HW) \emph{items}. Let us consider for instance an \ultrascale (ZCU102) platform \cite{ultrascale}: it is composed of several \arm cores, a \gpu and an \fpga.
If the ML model is spread over the different components, in particular on one \arm and the \fpga, there will necessarily be several items.
Indeed, 
the \arm associated code will be considered as one SW item
and will go through the
ED-12C/DO-178C \cite{do178c} development process.
Whereas the hardware design of the \fpga will be considered as another item (HW) and will go through the ED-80/DO-254 \cite{do254} development process.

\textbf{Requirement 3.}
Third, the implementation has to follow the usual aeronautical development standards
(e.g. ED-12C/DO-178C \cite{do178c} for software). 
Thus, the description of a model within the
format must be compliant with a certified implementation process.
This last objective concerns the capability to implement an item description following standards such as ED-12C/DO-178C \cite{do178c} or ED-80/DO-254 \cite{do254}.
Among the requirements from those standards, two are related to the format.
First, it must offer full traceability:
looking at the generated code (e.g. C or \cuda), it is humanly possible to trace back to the original exported
 model.
 Second, the execution must be \emph{predictable}, meaning that it is possible to assess a WCET (Worst Case Execution Time) \cite{WCETpaper}.
 This entails that the code is expected to be allocated statically on the resources, all the memory allocations are static and the schedule (here the sequence of operations) is also static.

\vspace{-.2cm}

\subsection{Contributions}
Our general objective is to define an approach  compatible with the ARP 6983 requirements presented in the prior section.
We focus on a representative subset of deep neural networks (DNNs)
that is feed-forward neural networks trained off-line.
Our main goal is
the definition of an adequate format, with a formal syntax and semantics,
able to describe both 1) a global ML model,
and 2) any parallelized allocation on several items,
the behaviour of which is equivalent to the global model.

\textbf{For requirement 1.}
There are several initiatives to propose an intermediate
format between trained models and their implementation such as
\onnx \cite{onnx} (Open Neural Network Exchange).
After a thorough evaluation of different existing formats,
we identified 
\nnef (Neural Network Exchange Format) \cite{nnefformat}
as the most suited for our purpose:
syntax and semantics are public
and moreover the authors made a strong effort to provide a formal specification. 

Since \nnef provided a potential candidate, we decided to construct our format on top of it.
As it is now, \nnef describes formally the global ML model. Indeed, the semantics of \nnef
is almost fully defined (see section \ref{sec-nnef}). What is however missing in \nnef
is the clear formalization of the \emph{execution model},
that is the formal behaviour behind a series of \emph{\nnef instructions}.
To fix this missing element, we rely on
\petri nets, a usual  representation of program behaviours \cite{peterson1977petri}.
This choice
is also consistent with the need to express distributed behaviour on items, as
\petri nets also allow to model all combinations of execution:
sequence, pipeline, recursion or parallel \cite{olderog1986operational}.

\textbf{For requirement 2.}
We illustrate in  section \ref{sec-distribution}
why decomposing ML model into items is of interest.
To that end, we extend the syntax and semantics of \nnef.
We rely in particular  on \emph{logical data exchange} among items to distribute the computation and express the semantics with \emph{coloured \petri nets}.
 We formally show that the synchronization of the \petri nets behaves as the \emph{original} non distributed \petri net.

\textbf{For requirement 3.}
 We do not propose a complete DO-178C compatible approach but
 instead show a reasonable
  implementation approach 
  on the \xavier platform \cite{nvidiajeston} 
  that we believe
 could be with some effort compliant with the ED-12C/DO-178C \cite{do178c}. 


%% file: currentpractice.tex
\begin{tikzpicture}[thick,scale=0.5]
\tikzstyle{block} = [draw,minimum height=0.1em, minimum width=.1cm, inner sep=3pt];
\tikzstyle{txt} = [text centered, inner sep=0pt];
\tikzstyle{compliance} = [dashed, draw,  inner sep=3pt ];

\draw (0,0) node[block] (TF) {Training framework};
\path (TF)+(0,-3) node [block, dashed] (ONNX) {\begin{tabular}{l}ML model description\\in intermediate format\end{tabular}};
\path (ONNX)+(0,-3) node [block] (TVM) {\begin{tabular}{l}Runtime ML model\\interpretor\end{tabular}};
\path (TVM)+(-2,-3) node [block] (cpu1) {CPU};
\path (cpu1)+(2.5,0) node [block] (cpu2) {CPU};
\path (cpu2)+(2.5,0) node [block] (ddr) {DDR};
\path (cpu1)+(2.4,-1) node [block, minimum width=3.5cm] (bus) {};
\path (cpu1)+(0,-2) node [block] (gpu) {GPU};
\path (gpu)+(2.5,0) node [block] (fpga) {FPGA};
\path (fpga)+(2.5,0) node [block] (dla) {DLA};

\node[
      draw,  
      inner sep=.2cm,
      fit={(cpu1) (cpu2) (ddr) (gpu) (fpga) (dla)},
      align=left,
       thick
    ] (Const) {};

\draw  [->]   (TF.south)  -- (ONNX.north)  ;
\draw  [->]   (ONNX.south)  -- (TVM.north) ;
\draw  [->,dashed]   (TVM.south) --($ (TVM.south) + (0,-.7)$)  -| (cpu1.north);
\draw  [->,dashed]    ($ (TVM.south) + (-0.5,0)$) --($ (TVM.south) + (-0.5,-.5)$)  -| ($ (gpu.west) + (-0.8,0)$) -- (gpu.west);
\draw  [->,dashed]    ($ (TVM.south) + (0.5,0)$) --($ (TVM.south) + (0.5,-.5)$)  -| ($ (dla.east) + (0.8,0)$)  -- (dla.east);

\draw (12,0) node[block] (TF) {Training framework};
\path (TF)+(0,-3) node [block, dashed] (nnef) {\begin{tabular}{l}ML model description\\in \emph{formal} format\end{tabular}};
\path (nnef)+(-3.5,-3) node [block, dashed] (nnef1) {i1 desc.};
\path (nnef)+(0,-3) node [block, dashed] (nnef2) {i2 desc.};
\path (nnef)+(3.5,-3) node [block, dashed] (nnef3) {i3 desc.};
\path (nnef1)+(0,-1.5) node [block, dashed] (exec1) {code};
\path (nnef2)+(0,-1.5) node [block, dashed] (exec2) {code};
\path (nnef3)+(0,-1.5) node [block, dashed] (exec3) {code};

\draw  [->]   (TF.south)  -- (nnef.north) ;
\draw  [->]   (nnef.south)  -- (nnef1.north) ;
\draw  [->]   (nnef.south)  -- (nnef2.north) ;
\draw  [->]   (nnef.south)  -- (nnef3.north) ;
\draw  [->]   (nnef1.south)  -- (exec1.north);
\draw  [->]   (nnef2.south)  -- (exec2.north);
\draw  [->]   (nnef3.south)  -- (exec3.north);

\path (exec2)+(-2.5,-2) node [block] (cpu1) {CPU};
\path (cpu1)+(2.5,0) node [block] (cpu2) {CPU};
\path (cpu2)+(2.5,0) node [block] (ddr) {DDR};
\path (cpu1)+(2.4,-1) node [block, minimum width=3.5cm] (bus) {};
\path (cpu1)+(0,-2) node [block] (gpu) {GPU};
\path (gpu)+(2.5,0) node [block] (fpga) {FPGA};
\path (fpga)+(2.5,0) node [block] (dla) {DLA};

\node[
      draw,  
      inner sep=.2cm,
      fit={(cpu1) (cpu2) (ddr) (gpu) (fpga) (dla)},
      align=left,
       thick
] (Const) {};

\draw  [->,dashed]   (exec1.south) -- (cpu1.north);
\draw  [->,dashed]    (exec1.west) --($ (exec1.west) + (-0.5,0)$)  -| ($ (gpu.west) + (-1.2,-0.1)$) -- ($ (gpu.west) + (0,-0.1)$);

\draw  [->,dashed]   (exec2.south) -- (cpu1.north);
\draw  [->,dashed]    ($ (exec2.south) + (-0.5,0)$) --($ (exec2.south) + (-0.5,-.3)$)  -| ($ (gpu.west) + (-.8,0.1)$) --  ($ (gpu.west) + (0,0.1)$);

\draw  [->,dashed]    (exec3.south) --($ (exec3.south) + (0,-.3)$)  -| ($ (dla.east) + (0.8,0)$) |-
($ (fpga.south) + (0,-0.8)$) --  (fpga.south);
\end{tikzpicture}

%% file: nnef.tex
\section{Format of description -- \nnef}
\label{sec-nnef}
\khronos standardization group\footnote{\url{https://www.khronos.org/}} has defined the \nnef (Neural Network Exchange format)
format with a specification that provides a syntax and a semantic.
We focus on the \nnef syntax and semantics elements needed for our purpose. The reader can refer
to \cite{nnefformat} for a complete description of \nnef.

\subsection{Brief Reminder on Neural Network}


The field of artificial intelligence has gained much research attention in the past years.
The power of AI resides in the capacity of solving highly complex problems \cite{Goodfellow-et-al-2016}. 
Machine learning domain describes the study and development of statistical algorithms that are able to efficiently generalize on unknown input data after the extraction of patterns from a similar, and representative, data set.
Neural networks are a class of ML algorithms. A neural network implements a mathematical function \Fn that aims at approximating a continuous real-valued function \cite{Hornik1989, Schafer2006}. 
\Fn is  composed of  different mathematical functions called \emph{layers}.

There are two types of deep neural networks: feed-forward neural networks and recurrent neural networks. 
Recurrent neural networks (RNN) feature layers that take as input some of their output (or the output of a successor layer), thus creating cycles. In feed-forward variants it is not true. We are not addressing RNN in this paper.
A common representation of a feed-forward deep neural network (FDNN) is in the form of a \emph{directed acyclic graph} (DAG) defining how its layers are connected together.

\begin{definition}[Feed-forward Deep Neural Network] \label{def:dnn}
  \it
  A \emph{feed-forward deep neural network} $\mathcal{N}=(V,E)$ is a directed acyclic graph, wherein:
  \begin{itemize}
  \item $V$ is the finite set of vertices of the graph, which represent its layers $l \in V$;
  \item $E\subseteq V\times V$ is the set of edges of the graph, representing the data flow within the neural network.
  \end{itemize} 
\end{definition}

In order to construct the possible flows of data within the feed-forward deep neural network, it is necessary to define what are the predecessors and successors of a vertex, or layer. 

\begin{definition}[Predecessors / successors of a layer] \label{def:pre-layer}
  \it
The direct predecessors (resp. successors) of a layer $l$ are defined as the layers of the set 
$Pre(l) = \{l' \in V \mid (l', l) \in E\}$ (resp.
$Succ(l) = \{l' \in V \mid (l, l') \in E\}$).
The predecessor transitive closure of $l$ is the set of all its predecessors layers noted
 $\emph{Pre}^*(l)=\bigcup_{n=1}^{k} \emph{Pre}^n(l)$, wherein:
 
    $\emph{Pre}^n(l) = 
\begin{cases}
    \emph{Pre}(l),& \text{if } n = 1\\
    \bigcup_{l' \in\emph{Pre}^{n-1}(l)} (\emph{Pre}(l')\medcup\{l'\}),              & \text{otherwise}
\end{cases}
$
\end{definition}

A layer can be classified into input, output and intermediate, or hidden.
An input layer $l$ only consumes input data, i.e., ${Pre(l) = \emptyset}$. Similarly, a final layer $l$ only produces output data, i.e., $Succ(l) = \emptyset$. We represent $V_I\subseteq V$ as the set of input layers and $V_O\subseteq V$ as the set of output layers. Note that $V_I\medcap V_O=\emptyset$. The remaining layers, $l\not \in V_I \medcup V_O$, are the hidden layers.
Finally, as a straight consequence of directed acyclic graphs properties, 
$l\not \in \emph{Pre}^*(l)$.



The function performed by a layer is of the form $f_l:\mathbb{R}^{m} \rightarrow \mathbb{R}^{n}$, wherein $m$ and $n$ represent respectively the input and output dimensions of the given layer's function. 
%
\begin{definition}[Function associated to a FDNN]
  \label{def:func-DNN}
  \it
  Let $\mathcal{N}=(V,E)$ be a feed-forward deep neural network and $V_O\subseteq V$ be the set of output layers. Let us denote the function associated to a set of layers such that:
\begin{equation}
\forall U \subseteq V, F_{U} =
\begin{cases}
(F_{l_{1}}, \ldots, F_{l_{n}}), & \text{if } U = \{l_1, ...,l_n\} \\
 F_{ \emptyset},                & \text{if } U = \emptyset.
\end{cases}
\end{equation}
wherein:
\begin{equation}
	 \forall l \in V, \quad F_{l}=f_{l}(F_{Pre(l)})   
\end{equation}
\end{definition}

We note $\Fn = F_{V_O}$ the function associated to a feed-forward deep neural network.

\begin{example}[\emph{Single-path} feed-forward deep neural network]
  \label{ex-lenet}
  \it
  It corresponds to the particular case of a feed-forward deep neural network, wherein:
 $V = \{l_1, \ldots, l_n\}$ and  $\forall i\geq 2$, $\emph{Pre}(l_i)=\{l_{i-1}\}$.
Therefore, $V_I=\{l_1\}$, $V_F=\{l_n\}$. 
Such an example is the \lenet shown in figure \ref{fig:lenet}.

\begin{figure}[hbt]
  \centering
  \resizebox{.9\linewidth}{!}{\input{lenet}}
  \caption{\lenet neural network\label{fig:lenet}}
\end{figure}

According to Definition \ref{def:func-DNN} the function associated to a single-path DNN
is the composition function:
$  F_\mathcal{N}(x) = \circ f_{l_{n-1}} \circ \ldots \circ f_{l_1}(x)$
wherein $m_{l_1}$ in the input dimension of $f_{l_1}$ and $p_{l_n}$ is the output dimension of $f_{l_n}$. 
\end{example}

\subsection{Neural Network Description in \nnef}
\label{sec-nn-nnef}
Our first goal concerns the definition of a format,
with a formal syntax and semantics, able to describe any ML model
such as the \lenet of example \ref{ex-lenet}.
Let us explain why \nnef fulfils this goal.
An \nnef description is composed of two parts:
1) A computation graph described in a human readable text file;
and 2) the parameters provided in multiple raw data file.
The fact that the description is textual is important for the traceability between the specification (output of the training framework) and the embedded code.

The computation graph file describes all parameters needed and operations to be done. More precisely, 
each line of the computation graph is an elementary instruction (\nnef \emph{compound fragment}) that may be split
into several atomic operations (\nnef \emph{primitives}).
The result of each instruction is stored in a named variable, that represents a tensor,
and which can be used as input for other instruction(s).
To compute an operation all its inputs variables shall be available.

\begin{remark}
  \nnef description follows a SSA (static single assignment) form \cite{cytron1989efficient}
  which helps the implementation process.
  It is usual to translate a program in its 
  SSA form before compilation or optimization passes \cite{BourkeBDLPR17}.
\end{remark}

\begin{example}
  \it
Let us illustrate how to specify a neural network with an example.
  The \nnef textual specification of the \lenet detailed in example \ref{ex-lenet}
  is given in the listing \ref{list-nnef-lenet}. First, all parameters are declared and stored as variables. \emph{e1} is the input tensor of size $1\times 32\times 32$,
  \emph{v1} contains the 6 kernels of size $1\times 5\times 5$ of the first convolution and
  \emph{v2} is the bias applied at the end of the  first convolution.
  The parameters needed by the layers should all be defined
  as variables in the description file. 

 \begin{figure}[hbt] 
\resizebox{1.3\textwidth}{!}{
\begin{lstlisting}[ caption={\lenet with \nnef syntax},
                  label=list-nnef-lenet]
graph torch_jit_export(e1) -> (out)
{
    e1 = external<scalar>(shape = [1, 1, 32, 32]);
    v1 = variable<scalar>(shape = [6, 1, 5, 5], label = 'v1');
    v2 = variable<scalar>(shape = [1, 6], label = 'v2');
    v3 = variable<scalar>(shape = [16, 6, 5, 5], label = 'v3');
    v4 = ...; v5 = ...; v6 = ...; v7 = ...;
    v8 = ...; v9 = ...; v10 = ...;
    
    o1 = conv(e1, v1, v2, stride = [1, 1], dilation = [1, 1], padding = [(0, 0), (0, 0)], groups = 1);
    o2 = relu(o1);
    o3 = max_pool(o2, size = [1, 1, 2, 2], stride = [1, 1, 2, 2], dilation = [1, 1, 1, 1], padding = [(0, 0), (0, 0), (0, 0), (0, 0)], border = 'ignore');
    o4 = conv(o3, v3, v4, ...;   o5 = relu(o4);
    o6 = max_pool(o5, ...;  o7 = reshape(o6, shape = [0, -1]);
    o8 = linear(o7, v5, v6);  o9 = relu(o8);
    o10 = linear(o9, v7, v8);  o11 = relu(o10);
    o12 = linear(o11, v9, v10);
    out = softmax(o12, axes = [1]);
}
\end{lstlisting}}
\end{figure}

 After the variables declaration, comes the computation graph itself.
 The output of the first convolution is stored in the variable \emph{o1}.
 When calling the function / compound fragment \emph{conv}, the user must instantiate
 the full set of parameters for this type of layer:
 input tensor, kernels, bias,  stride, dilation, padding and groups.
 Every parameter appears explicitly in the definition and there is no ambiguity.
 For instance,
 the way to declare the padding explicitly states how the padding applies on top / bottom / right / left.
 After the convolution, the activation function has to be explicitly applied to \emph{o1}
 and thus is not hidden in the convolution, ensuring again an unambiguous description.
 The first pooling layer results in variable \emph{o3}.
 Reading the description, we recognize the \lenet detailed before. 
 The flat layer is encoded with a more expressive function \emph{reshape} that allows several reshaping.
 The \emph{dense} layer is called \emph{linear}. The post-processing \emph{softmax} is also explicit.
\end{example}

The \nnef specification also
provides the link between instructions (e.g. \emph{conv}) appearing in the file
and their associated mathematical functions.
Subsequently,
we will not use \nnef terms, because the \nnef standard uses the generic term \emph{fragment} for both. 
We illustrate this with the max pooling layer only.

\subsection{Max Pooling Layer Semantics}
\label{sec:maxpool}
Let us illustrate issues that may arise without a formal description.
Let us first remind  the functional semantics of a max pooling
layer where a padding (and no dilatation) is applied.
Thus, the function is defined by $\mathcal{P}\emph{ool}_{k,s} \circ \mathcal{P}_{p,v}$
where each function is defined below.
\begin{definition}[Padding associated function -- $\mathcal{P}_{p,v}$]
  \label{def-pad}
  \it
  Let  $p=(p_t,p_b,p_l,p_r)$ be a 4-tuple of integers representing the padding to be applied on each border of a 3D-tensor and $v$ the float value to be used for the padding.
  The padding function  $\mathcal{P}_{p,v}$ applied on a 3D-tensor $I$ of size $(n_h,n_w,n_c)$
  outputs a 3D-tensor $O = \mathcal{P}_{p,v}(I)$ of size $(o_h,o_w,o_c)$ with
  $o_h = n_h + p_t + p_b$, $o_w = n_w + p_l + p_r$ and $o_c = n_c$ such that\\
  $O_{x,y,z}=
    \begin{cases}
            v  & \textbf{if }( x \leq p_t)  \emph{ or } (x > n_h + p_t)  \emph{ or }\\
            &   \ \ \   (y \leq p_l)  \emph{ or } (y > n_w + p_l)\\
            I_{x-p_t,y-p_l,z}  &\textbf{otherwise}
    \end{cases}
  $
\end{definition}

\begin{definition}[Pooling layer associated function --  $\mathcal{P}\emph{ool}_{k,s}$]
  Let $s=(s_h,s_w)$ be the \emph{stride} parameters and
   $k=(k_h,k_w)$ be the height and width of the \emph{window}.
  The pooling applied on a 3D-tensor $I$ of size $(n_h,n_w,n_c)$
  outputs a 3D-tensor $O=\mathcal{P}\emph{ool}_{k,s}(I)$ of size $(o_h,o_w,o_c)$
  with $o_h = \left\lfloor \frac{n_h - k_h}{s_h} + 1\right\rfloor$,
  $o_w = \left\lfloor \frac{n_w - k_w}{s_w} + 1 \right\rfloor$ and
  $o_c = n_c$ with
  $O_{x,y,z} = \emph{max}(I[s_h \cdot (x-1)+1:s_h \cdot (x-1)+k_h+1][s_w \cdot (y-1)+1:s_w \cdot (y-1)+k_w+1][z])$.
  Here, $I[s_{11}:s_{21},...,s_{1k}:s_{2k}]$ represents the \emph{slice} of $I$
of all the values $I_{s_{11}+x_1,...,s_{1k}+x_k}$ with $i \in [1, k]$ and $x_i \in [1,s_{2i}-s_{1i}]$.
 \end{definition}

The \nnef syntax of a max pooling layer describes the padding values with an enumerate string. Ignoring the border for a max pooling layer is equivalent to pad with the minimum float value (\emph{MIN_F}). This corresponds to the neutral operator of the max function.
 Looking now at the \emph{max_pool} elementary instruction according to \nnef documentation \cite{nnefformat}, it is defined in the pseudo-code
  by 2 atomic operations. 
  \begin{enumerate*}
    \item \emph{argmax_pool} that computes an array of index (corresponding to the max in each pool);
    \item \emph{sample} that returns for an array of index, an array with corresponding values.
  \end{enumerate*}
  This pseudo-code indeed encodes the expected function $\mathcal{P}\emph{ool}_{k,s} \circ \mathcal{P}_{p,v}$.
  
 \textbf{Of the importance of unambiguous description.} 
  We propose to highlight the importance of making unambiguous textual descriptions of ML models
  by comparing two state-of-the-art training frameworks, namely \pytorch and \keras (with \tensorflow).
\begin{itemize}
  \item In  \keras, a padding inside a max pooling layer can only be declared by a string $\in \{$"valid", "same"$\}$.
    "valid" means no elements to add,
    while "same" means that padding is added on the right and on the bottom borders only to fit  the size of the pool.

    \begin{lstlisting}[ %caption={\keras syntax for MaxPool},
      label=list-torch, language=Python]
  # KERAS SYNTAX for pool1
      MaxPool = tf.keras.layers.MaxPooling2D(pool_size=(2,2),strides=(2,2),padding = 'same')

  # NNEF for KERAS SYNTAX
      max_pool_keras = max_pool(input, size = [1, 1, 2, 2], stride = [1, 1, 2, 2], dilation = [1, 1, 1, 1], padding = [(0, 0), (0, 0), (0, 1), (0, 1)], border = 'ignore');
    \end{lstlisting}
This corresponds to
$  \mathcal{P}ool_{[2,2],[2,2]} \circ \mathcal{P}_{[0,1,0,1],\textrm{MIN\_F}}$.
\item In
  \pytorch, a padding inside a max pooling layer can only be declared by one or two integers. In case of one integer, this defines the number of elements to add to each border (top, bottom, left and right). In presence of 2 integers,
  the first  gives the number of elements to add at the top and  bottom, and the second at the left and right.
   \begin{lstlisting}[ %caption={\pytorch syntax for MaxPool},
      label=list-torch, language=Python]
  #PYTORCH SYNTAX for pool1
     MaxPool = NN.MaxPool2d(2,              2,       1)
                         # Kernel Size,  Stride, Padding

  #NNEF for PYTORCH SYNTAX
     max_pool_torch = max_pool(input, size = [1, 1, 2, 2], stride = [1, 1, 2, 2], dilation = [1, 1, 1, 1], padding = [(0, 0), (0, 0), (1, 1), (1, 1)], border = 'ignore');
   \end{lstlisting}
 This corresponds to $\mathcal{P}ool_{[2,2],[2,2]} \circ \mathcal{P}_{[1,1,1,1],\textrm{MIN\_F}}$
\end{itemize}
The semantics of \keras and \pytorch are not equivalent,
and there is no possibility to convert one into another at once.
The only way to make a valid conversion is  to explicitly add a padding layer before the max pooling.

\subsection{\nnef Execution Model Semantics}
\label{subsec-petrinet-sem}
The semantics of the \emph{execution model}, that is the formal behaviour behind a series of \emph{\nnef instructions},
is not explicitly given by the standard. It assumes that one instruction can be executed
when all its inputs are computed and available.
Thus, executing the instructions in sequence following
the order of the \nnef guarantees a correct execution.
There are two types of instructions: those reading parameters from binary files or input tensors
and  layer-associated instructions based on one or several atomic operations.
An instruction is always of the form
\[
 \emph{var} = \emph{operation} (v_1, \ldots, v_k, \emph{cst}_1,\ldots, \emph{cst}_j);
 \]
 where \emph{var} is a variable computed by the \emph{operation} (any \nnef \emph{fragment}),
 $v_i$ are either variables computed beforehand, the input tensor or fixed parameters (e.g. kernels), and
 \emph{cst}$_i$ are constant (e.g. stride).
 The \nnef execution model can be formally expressed using the \petri net formalism \cite{peterson1977petri}.
 
\begin{translation}
  \label{trans-nnef-petri}
A \nnef description, composed of $n$ instructions, generates a \petri net $(P,T,V)$
where:
\begin{itemize}
\item the set of places $P$ corresponds to all variables appearing in the \nnef description
  (i.e. $n$ places for a description of $n$ instructions);
  \begin{itemize}
  \item a token in a place means that the associated variable is available for computation;
  \item initially there are as many tokens in each parameter-associated place as the parameter is needed in the instructions
    and as many tokens in the input tensor place as the input tensor is used by the instructions;
    \item there is a unique final place corresponding to the last variable computed in the \nnef file;
\end{itemize}
\item the set of transition names $T$ corresponds to all instructions appearing in the \nnef description;
\item $V\subseteq 2^P \times T \times \mathbb{N} \times P$ defines the set of transitions.
  A transition can be fired if there is a token in all input places.
  When it fires, the transition removes a token from each of these places and generates as many tokens as defined on the edge in the unique output place.
  \begin{itemize}
  \item each instruction $\emph{var} = \emph{op} (v_1, \ldots, v_k,\emph{cst}_1,\ldots, \emph{cst}_j)$ generates the transition given in figure \ref{fig:trans-instruction}
    where $p$ is the number of time \emph{var} will be consumed by other instructions;
    \item when only one token is produced by a transition, we omit the integer.
    \end{itemize}
  \end{itemize}
\end{translation}

\begin{figure}[hbt]
  \centering
  \hspace*{\fill}
  \begin{minipage}{0.22\linewidth}
  \resizebox{\linewidth}{!}{\input{petri-net-trans}}
  \caption{Translation of one instruction\label{fig:trans-instruction}}
  \end{minipage}
  \hspace*{\fill}
  \begin{minipage}{0.70\linewidth}
  \resizebox{\linewidth}{!}{\input{petri-net}}
  \caption{\nnef semantics of the \lenet express with \petri net. Initial marking\label{fig:sem}}
  \end{minipage}
  \hspace*{\fill}
\end{figure}

The semantic of the \petri net clarifies the execution order that is unclear in the \nnef formalism.
Especially, the order of the textual file should not impose a unique order when several valid ones may exist.
\begin{example}
  \it
  The \lenet model  of figure \ref{fig:lenet} with its associated \nnef description in listing \ref{list-nnef-lenet} has the associated \petri net given in figure \ref{fig:sem}. We recognize the instructions sequence that describes the computation of the neural network graph.
  There is a unique possible schedule for this \nnef description,
  but we will see later other \nnef models that accept several schedules (see section \ref{sec-distribution}).
\end{example}

\begin{remark}
  The \petri net of figure \ref{fig:sem} only defines the semantics of a single inference pass. It is usual to repeat the inference pass
  in order to process new inputs (e.g. in a periodic manner).
  This can also be represented using the \petri formalism by sending back tokens to $e1$ and $v_k$ places.
  For demonstration and clarity, subsequently in the paper, we always consider the semantics of a single inference.
\end{remark}

A way to define the semantics of \petri nets is to compute the set of reachable \emph{markings}, where a marking
 defines the number of tokens in each place at a given instant.
 \begin{definition}[Marking]
   \it
  Considering a \petri net with $n$ places, a marking is defined as a vector $v\in\mathbb{N}^n$
  giving the number of tokens $v[i]$ in the i-th place.
 This initial location of tokens is called the \emph{initial marking},
 this represents the starting state of the system.
 A \emph{final marking} is a marking such that there is one token
 only in each final place
 and from which no transition can be fired any more.
\end{definition}
 \begin{example}
   \it
  In the \petri net of figure \ref{fig:sem}, there are 24 states. There are 11 tokens in the initial making and there is a single token in the unique final marking
  (in the place \emph{out}).
\end{example}

\begin{property}[Initial marking and unique final marking]
  The unique initial marking is defined by the translation
  and consists of token(s) in the input tensor variable place and parameters-associated places.
  Because we consider feed-forward neural networks,
  there is unique \emph{final marking}.
\end{property}

\begin{definition}[Paths and semantics]
  \it
  A \emph{valid path} starts from the initial marking $m_i$,
  lists a series of fireable transitions 
  and ends in a final marking $m_f$, i.e.
    $m_i \longrightarrow^{t_1} m_1 \longrightarrow^{t_2} \ldots \longrightarrow^{t_l} m_f$.
  The semantics of a \petri net is the set of valid paths.
\end{definition}

\begin{property}[Possible executions of an \nnef description]
  Because we consider feed forward neural networks, the number of valid paths is finite and
  the valid paths correspond to all possible execution orders
  respecting the semantics of the ML model.
\end{property}
Semantics preserving code generation could lead to
any implementation the path of which is valid. Full sequential code following the order of instructions of the \nnef file is one of them.

%% file: lenet.tex
\tikzstyle{txt} = [text centered, inner sep=0pt]

\begin{tikzpicture}[scale=1.0,
z={({0.3cm*cos(45)},{0.3cm*sin(45)})},
>=latex, 
font=\footnotesize,
]


\draw[fill=white] (0.25+7*0.05,7*0.05) rectangle (1+7*0.05,.75+7*0.05);
\foreach \foreach \z in {2,...,0}  {%
 \draw[fill=white] (0.25+\z*0.05,.1+\z*0.05) rectangle (1+\z*0.05,.85+\z*0.05);
}
\draw (0.7,1.5)node[txt] (conv1){conv1};
\draw[->] (-1,0.5) -- node[above] {$1$x$32$x$32$} (0.25,0.5);
\draw[<->] (.25-0.2,.85+0*0.05) -- node[left] {$6$} (.25+0.3,.85+0.5);

\draw[fill=white] (2.5+5*0.05,5*0.05) rectangle (3.25+5*0.05,0.85+5*0.05);
\foreach \foreach \z in {2,...,0}  {%
 \draw[fill=white] (2.5+\z*0.05,0.1+\z*0.05) rectangle (3.25+\z*0.05,.85+\z*0.05);
}
\draw (3.2,1.5)node[txt] (conv1){pool1};
\draw[->] (1,0.5) -- node[above,pos=0.6] {$6$x$28$x$28$} (2.5,0.5);
\draw[<->] (2.5-0.2,0.85+0*0.05) -- node[left] {$6$} (2.5+0.3,0.85+0.5);

\draw[fill=white] (4.75+7*0.05,7*0.05) rectangle (5.5+7*0.05,0.75+7*0.05);
\draw (5.5,1.5)node[txt] (conv1){conv2};
\draw[->] (3.25,0.5) -- node[above,pos=0.6] {$6$x$14$x$14$} (4.75,0.5);
\foreach \foreach \z in {2,...,0}  {%
 \draw[fill=white] (4.75+\z*0.05,0.1+\z*0.05) rectangle (5.5+\z*0.05,0.85+\z*0.05);
}
\draw[<->] (4.75-0.2,0.85+0*0.05) -- node[left] {$16$} (4.75+0.3,0.85+0.5);

\draw[fill=white] (7+7*0.05,7*0.05) rectangle (7.75+7*0.05,0.75+7*0.05);
\draw[->] (5.5,0.5) -- node[above,pos=0.67] {$16$x$10$x$10$} (7,0.5);
\foreach \foreach \z in {2,...,0}  {%
 \draw[fill=white] (7+\z*0.05,0.1+\z*0.05) rectangle (7.75+\z*0.05,0.75+\z*0.05);
}
\draw (7.75,1.5)node[txt] (conv1){pool2};
\draw[<->] (7-0.2,0.85+0*0.05) -- node[left] {$16$} (7+0.3,0.85+0.5);
\draw[->] (7.75,0.5) -- node[above,pos=0.6] {$16$x$5$x$5$} (9,0.5);

\draw[fill=white] (9,0) rectangle (9.3,1);
\draw[->] (9.3,0.5) -- node[above] {$400$} (9.8,0.5);
\draw (9,1.5)node[txt] (conv1){flat};

\draw[fill=white] (9.8,0) rectangle (10.1,1);
\draw[fill=white] (10.6,0) rectangle (10.9,1);
\draw[fill=white] (11.4,0) rectangle (11.7,1);
\draw[->] (10.1,0.5) --node[above] {$120$}(10.6,0.5);
\draw[->] (10.9,0.5) --node[above] {$84$} (11.4,0.5);
\draw (10.5,1.5)node[txt] (conv1){dense1 \hspace{.1cm} 2 \hspace{.5cm} 3};

\draw[->] (11.7,0.5) -- node[above] {$10$} (12,0.5);

\end{tikzpicture}

%% file: petri-net-trans.tex
\begin{tikzpicture}[
thick,
node distance=1cm,
on grid,
post/.style={-{Latex[length=1.5mm]}},
every transition/.style={fill,minimum width=1mm, minimum height=10mm}
]

\node[place,tokens=0] (place1) {$v_2$};
\node[place,tokens=0,above=1cm of place1] (place3) {$v_1$};
\node[place,tokens=0,below=1.5cm of place1] (place4) {$v_k$};
\path (place1) +(0,-.75) node {\ldots};

\node[transition,right=1cm of place1,label=below:op] (trans1) {};
\node[place,tokens=0,right=1cm of trans1] (place2) {var};

\draw (place1) edge[post] (trans1);
\draw (place3) edge[post] (trans1);
\draw (place4) edge[post] (trans1);
\draw (trans1) edge[post] (place2) node[below right] {p};
\end{tikzpicture}

%% file: petri-net.tex
\begin{tikzpicture}[
thick,
node distance=1.1cm,
on grid,
post/.style={-{Latex[length=1.5mm]}},
every transition/.style={fill,minimum width=1mm, minimum height=10mm}
]

\node[place,
    tokens=1, label=below:$e_1$] (place1) {};

\node[transition,right=0.75cm of place1,label=below:$Conv$] (trans1) {};
\node[place,tokens=0,right=0.75cm of trans1, label=below:$o_1$] (place2) {};

\node[place,tokens=1,above left=1.3cm and .1cm of place1, label= right:$v_1$] (placevar1) {};

\node[place,tokens=1,above=1cm of placevar1, label= right:$v_2$] (placevar3) {};

\node[transition,right=.75cm of place2,label=$ReLU$] (trans2) {};
\node[place,tokens=0,right=.75cm of trans2, label=below:$o_2$] (place3) {};

\node[transition,right=.75cm of place3,label=$MaxPool$] (trans3) {};
\node[place,tokens=0,right=.75cm of trans3, label=below:$o_3$] (place4) {};

\node[transition,right=.75cm of place4,label=below:$Conv$] (trans4) {};
\node[place,tokens=0,right=.75cm of trans4, label=below:$o_4$] (place5) {};

\node[place,tokens=1,above left=1.3cm and .5cm of place4, label=left:$v_3$] (placevar10) {};

\node[place,tokens=1,above=1.1cm of placevar10, label=left:$v_4$] (placevar12) {};

\node[transition,right=.75cm of place5,label=$ReLU$] (trans5) {};
\node[place,tokens=0,right=.75cm of trans5, label=below:$o_5$] (place6) {};

\node[transition,right=.75cm of place6,label=$MaxPool$] (trans6) {};
\node[place,tokens=0,right=of trans6, label=below:$o_6$] (place7) {};

\node[transition,right=.75cm of place7,label=$Flat$] (trans7) {};
\node[place,tokens=0,below=2cm of trans7.north, label=above:$o_7$] (place8) {};

\node[transition,left=.75cm of place8,label=$Dense$] (trans8) {};
\node[place,tokens=0,left=.75cm of trans8, label=above:$o_8$] (place9) {};

\node[place,tokens=1,below right=1.2cm and .25cm of place8, label=left:$v_5$] (placevar14) {};

\node[place,tokens=1,below=1cm of placevar14, label=left:$v_6$] (placevar16) {};

\node[transition,left=.75cm of place9,label=below:$ReLU$] (trans9) {};
\node[place,tokens=0,left=.75cm of trans9, label=above:$o_9$] (place10) {};

\node[transition,left=.75cm of place10,label=below:$Dense$] (trans10) {};
\node[place,tokens=0,left=.75cm of trans10,  label=above right:$o_{10}$] (place11) {};

\node[transition,left=.75cm of place11,label=below:$ReLU$] (trans11) {};
\node[place,tokens=0,left=.75cm of trans11, label=above:$o_{11}$] (place12) {};

\node[transition,left=.75cm of place12,label=below:$Dense$] (trans12) {};
\node[place,tokens=0,left=.75cm of trans12,  label=above:$o_{12}$] (place13) {};

\node[transition,left=.75cm of place13,label=below:$SoftMax$] (trans13) {};
\node[place,tokens=0,left=of trans13, label=above:$out$] (place14) {};

\node[place,tokens=1,below right=1.2cm and .5cm of place10, label= right:$v_{7}$] (placevar20) {};

\node[place,tokens=1,below=1cm of placevar20, label= right:$v_{8}$] (placevar22) {};

\node[place,tokens=1,below right=1.2cm and .5cm of place12, label= right:$v_{9}$] (placevar30) {};

\node[place,tokens=1,below=1cm of placevar30, label= right:$v_{10}$] (placevar32) {};

\draw (place1) edge[post] (trans1);
\draw (trans1) edge[post] (place2);
\draw (place2) edge[post] (trans2);
\draw (trans2) edge[post] (place3);
\draw (place3) edge[post] (trans3);
\draw (trans3) edge[post] (place4);
\draw (place4) edge[post] (trans4);
\draw (trans4) edge[post] (place5);
\draw (place5) edge[post] (trans5);
\draw (trans5) edge[post] (place6);
\draw (place6) edge[post] (trans6);
\draw (trans6) edge[post] (place7);
\draw (place7) edge[post] (trans7);
\draw (trans7) edge[post, bend left = 80] (place8);
\draw (place8) edge[post] (trans8);
\draw (trans8) edge[post] (place9);
\draw (place9) edge[post] (trans9);
\draw (trans9) edge[post] (place10);
\draw (place10) edge[post] (trans10);
\draw (trans10) edge[post] (place11);
\draw (place11) edge[post] (trans11);
\draw (trans11) edge[post] (place12);
\draw (place12) edge[post] (trans12);
\draw (trans12) edge[post] (place13);
\draw (place13) edge[post] (trans13);
\draw (trans13) edge[post] (place14);

\draw (placevar1) edge[post, bend right=5] (trans1);

\draw (placevar3) edge[post, bend left=5] (trans1);

\draw (placevar12) edge[post,bend right=5] (trans4);

\draw (placevar10) edge[post,bend right=15] (trans4);

\draw (placevar14) edge[post,bend right=15] (trans8);

\draw (placevar16) edge[post,bend right=5] (trans8);

\draw (placevar20) edge[post,bend right=15] (trans10);

\draw (placevar22) edge[post,bend right=5] (trans10);

\draw (placevar30) edge[post,bend right=15] (trans12);

\draw (placevar32) edge[post,bend right=5] (trans12);

\end{tikzpicture}

%% file: distribution.tex
\section{Is distribution needed?}
\label{sec-distribution}
Because we consider highly distributed platforms, a designer may choose to split the ML model into several parts
in order to accelerate the execution and reduce the execution time.
In particular, it could lead to developing parts independently and on different \emph{items} following the aeronautics standards \cite{arp4754}.
In such a case,  there should be a formal description for each item that becomes the input
specification for HW/SW item implementation. 
We identified 3 different needs for distribution to be addressed
that we illustrate on the \lenet example.

\begin{remark}
Note that if the designer considers its platform as a unique item, the \nnef description will be the specification.
\end{remark}

\noindent \textbf{Off-loading computation.}
  Let us consider for instance the implementation of the \lenet
  on an \ultrascale (ZCU102) platform \cite{ultrascale}
composed of several \arm cores, a \gpu and an \fpga.
Let us assume that we choose to execute the convolution on the \fpga and all other layers on one \arm.
The idea will be to offload the input tensor of each convolution on the \fpga
and retrieve the feature maps from the \fpga (see figure below).

\begin{figure}[hbt]
  \centering
  \resizebox{.95\linewidth}{!}{\input{lenet-fpga}}
\end{figure}


\noindent \textbf{Parallelizing the layers.}
A second type of distribution could be to refine the layers and exhibit more parallelism by distributing the computation of a layer across several items.
It will be up to the designer to show the semantic preservation at this refinement level.
  Looking again at the \lenet, we can split the computation of the first two layers on two different items. Each item
  will do the convolution+maxpool on a part of the input image.
  To do so, the input image is split along the height
  and two sub-parts are executed on two different items (see figure below).
  In order to keep the semantics for the convolution, some overlap exists between the two sub-images.

\begin{figure}[hbt]
  \centering
  \resizebox{.95\linewidth}{!}{\input{lenet-parallel}}

\end{figure}

\noindent \textbf{Pipelining the computation.}
  A third type of distribution is the pipelining of the DNN execution.
  In this case, each item is in charge of computing a specific layer (or a group of layers).
  The first item computes the first layer(s) on the first tensor input and sends its output to the second item that will compute the second group of layer(s)
  while the first item starts processing the second input tensor. 
  This classical mechanism enables to reduce (e.g. for video processing) the frame rate while degrading the latency. 
  The depth of the pipeline is the number of inputs that can be handled at the same time among the pipeline.

  \begin{figure}[hbt]
    \centering
    \resizebox{.95\linewidth}{!}{\input{lenet-pipeline}}
  \end{figure}

%% file: lenet-fpga.tex
\tikzstyle{txt} = [text centered, inner sep=0pt]

\begin{tikzpicture}[scale=1.0,
z={({0.3cm*cos(45)},{0.3cm*sin(45)})},
>=latex, 
font=\footnotesize,
]


\draw[fill=white] (0.25+7*0.05,7*0.05) rectangle (1+7*0.05,.75+7*0.05);
\foreach \foreach \z in {2,...,0}  {%
 \draw[fill=white] (0.25+\z*0.05,.1+\z*0.05) rectangle (1+\z*0.05,.85+\z*0.05);
}
\draw (0.9,1.3)node[txt] (conv1){conv1};
\draw[->] (-.8,0.5) -- node[above] {$1$x$32$x$32$} (0.25,0.5);
\draw[<->] (.25-0.2,.85+0*0.05) -- node[left] {$6$} (.25+0.3,.85+0.5);

\draw[fill=white] (2.5+5*0.05-.7,5*0.05+1.5) rectangle (3.25+5*0.05-.7,0.85+5*0.05+1.5);
\foreach \foreach \z in {2,...,0}  {%
 \draw[fill=white] (2.5+\z*0.05-.7,0.1+\z*0.05+1.5) rectangle (3.25+\z*0.05-.7,.85+\z*0.05+1.5);
}
\draw (3.2-.7,1.5+1.5)node[txt] (conv1){pool1};
\draw[->] (1,0.5) --(1.75-.4,0.5) |- node[above,pos=0.6] {$6$x$28$x$28$} (2.5-.7,0.5+1.5);
\draw[<->] (2.5-0.2-.7,0.85+0*0.05+1.5) -- node[left] {$6$} (2.5+0.3-.7,0.85+0.5+1.5);

\draw[fill=white] (4.75+7*0.05-1.2,7*0.05) rectangle (5.5+7*0.05-1.2,0.75+7*0.05);
\draw (5.5-1.2,1.3)node[txt] (conv1){conv2};
\draw[->] (3.25-.7,0.5+1.5) -- (4-1,0.5+1.5) |- node[above,pos=0.6] {$6$x$14$x$14$} (4.75-1.2,0.5);
\foreach \foreach \z in {2,...,0}  {%
 \draw[fill=white] (4.75+\z*0.05-1.2,0.1+\z*0.05) rectangle (5.5+\z*0.05-1.2,0.85+\z*0.05);
}
\draw[<->] (4.75-0.2-1.2,0.85+0*0.05) -- node[left] {$16$} (4.75+0.3-1.2,0.85+0.5);

\draw[fill=white] (7+7*0.05-1.4,7*0.05+1.5) rectangle (7.75+7*0.05-1.4,0.75+7*0.05+1.5);
\draw[->] (5.5-1.2,0.5) -- (6.25-1.4,0.5) |- node[above,pos=0.67] {$16$x$10$x$10$} (7-1.4,0.5+1.5);
\foreach \foreach \z in {2,...,0}  {%
 \draw[fill=white] (7+\z*0.05-1.4,0.1+\z*0.05+1.5) rectangle (7.75+\z*0.05-1.4,0.75+\z*0.05+1.5);
}
\draw (7.75-1.4,1.5+1.5)node[txt] (conv1){pool2};
\draw[<->] (7-0.2-1.4,0.85+0*0.05+1.5) -- node[left] {$16$} (7+0.3-1.4,0.85+0.5+1.5);
\draw[->] (7.75-1.4,0.5+1.5) -- node[above,pos=0.6] {$16$x$5$x$5$} (9-1.4,0.5+1.5);

\draw[fill=white] (9-1.4,0+1.5) rectangle (9.3-1.4,1+1.5);
\draw[->] (9.3-1.4,0.5+1.5) -- node[above] {$400$} (9.8-1.4,0.5+1.5);
\draw (9-1.4,1.5+1.5)node[txt] (conv1){flat};

\draw[fill=white] (9.8-1.4,0+1.5) rectangle (10.1-1.4,1+1.5);
\draw[fill=white] (10.6-1.4,0+1.5) rectangle (10.9-1.4,1+1.5);
\draw[fill=white] (11.4-1.4,0+1.5) rectangle (11.7-1.4,1+1.5);
\draw[->] (10.1-1.4,0.5+1.5) --node[above] {$120$}(10.6-1.4,0.5+1.5);
\draw[->] (10.9-1.4,0.5+1.5) --node[above] {$84$} (11.4-1.4,0.5+1.5);
\draw (10.5-1.4,1.5+1.5)node[txt] (conv1){dense1 \hspace{.1cm} 2 \hspace{.5cm} 3};

\draw[->] (11.7-1.4,0.5+1.5) -- node[above] {$10$} (12-1.4,0.5+1.5);

\coordinate (item1A) at (-.8 , 0);
\coordinate (item1B) at (10.4,1.4);       
\draw [fill=gray!40, opacity=.15](item1A) rectangle (item1B) 
                    node[xshift=-3.5ex, yshift=-2.2ex, opacity=1] {\begin{tabular}{l}item1\\\fpga\end{tabular}};

\coordinate (item2A) at (-.8 , 1.5);
\coordinate (item2B) at (10.4,3.2); 
 \draw [fill=gray!200, opacity=.15](item2A) rectangle (item2B) 
                    node[xshift=-10cm, yshift=-2.5ex, opacity=1] {\begin{tabular}{l}item2\\\arm\end{tabular}};
\end{tikzpicture}

%% file: lenet-parallel.tex
\tikzstyle{txt} = [text centered, inner sep=0pt]

\begin{tikzpicture}[scale=1.0,
z={({0.3cm*cos(45)},{0.3cm*sin(45)})},
>=latex, 
font=\footnotesize,
]


\draw[->] (-3,1.5) -- node[above] {$1$x$32$x$32$} (-2,1.5);

\draw[fill=white] (-2,1) rectangle (-1.5,2);
\draw (-1.75,2.25)node[txt] (conv1){split};
\draw[-] (-1.5,1.7) -- (-1.2,1.7) -| (-1.2,2.5);
\draw[-] (-1.5,1.3) -- (-1.2,1.3) -| (-1.2,.25);

\draw[fill=white] (0.25+4*0.05-.6,4*0.05) rectangle (1+4*0.05-.6,.4+4*0.05);
\foreach \foreach \z in {2,...,0}  {%
 \draw[fill=white] (0.25+\z*0.05-.6,.1+\z*0.05) rectangle (1+\z*0.05-.6,.4+\z*0.05);
}
\draw (0.7-.6,1)node[txt] (conv1){conv1};
\draw[->] (-1.2,0.25) -- node[above] {$1$x$18$x$32$} (0.25-.6,0.25);
\draw[<->] (.25-0.2-.6,.4+0*0.05) -- node[left] {$6$} (.25+0.4-.6,.4+0.5);

\draw[fill=white] (2.5+4*0.05-.9,4*0.05) rectangle (3.25+4*0.05-.9,0.40+4*0.05);
\foreach \foreach \z in {2,...,0}  {%
 \draw[fill=white] (2.5+\z*0.05-.9,0.1+\z*0.05) rectangle (3.25+\z*0.05-.9,.40+\z*0.05);
}
\draw (3.2-.9,1)node[txt] (conv1){pool1};
\draw[->] (1-.6,0.25) -- node[above,pos=0.6] {$6$x$14$x$28$} (2.5-.9,0.25);
\draw[<->] (2.5-0.2-.9,0.4+0*0.05) -- node[left] {$6$} (2.5+0.3-.9,0.4+0.5);

\draw[fill=white] (-2,1) rectangle (-1.5,2);
\draw (-1.75,2.25)node[txt] (conv1){split};

\draw[fill=white] (0.25+4*0.05-.6,4*0.05+2.25) rectangle (1+4*0.05-.6,.40+4*0.05+2.25);
\foreach \foreach \z in {2,...,0}  {%
 \draw[fill=white] (0.25+\z*0.05-.6,.1+\z*0.05+2.25) rectangle (1+\z*0.05-.6,.40+\z*0.05+2.25);
}
\draw (0.7-.6,1+2.25)node[txt] (conv12){conv1};
\draw[->] (-1.2,2.5) -- node[above] {$1$x$18$x$32$} (0.25-.6,2.5);
\draw[<->] (.25-0.2-.6,.4+0*0.05+2.25) -- node[left] {$6$} (.25+0.3-.6,.4+0.5+2.25);

\draw[fill=white] (2.5+5*0.05-.9,5*0.05+2.25) rectangle (3.25+5*0.05-.9,0.40+5*0.05+2.25);
\foreach \foreach \z in {2,...,0}  {%
 \draw[fill=white] (2.5+\z*0.05-.9,0.1+\z*0.05+2.25) rectangle (3.25+\z*0.05-.9,.40+\z*0.05+2.25);
}
\draw (3.2-.9,1+2)node[txt] (conv12){pool1};
\draw[->] (1-.6,0.5+2) -- node[above,pos=0.6] {$6$x$14$x$28$} (2.5-.9,0.5+2);
\draw[<->] (2.5-0.2-.9,0.4+0*0.05+2.25) -- node[left] {$6$} (2.5+0.3-.9,0.4+0.5+2.25);

\draw[fill=white] (4.75-1.3,1.1) rectangle (5.0-1.3,1.85);
\draw (4.80-1.1,2)node[txt] (concat){concat};
\draw[->] (3.25-.9,0.25) -| node[above,pos=0.6] {$6$x$7$x$14$} (4-1.3,0.5+.8) -- (4.75-1.3,0.5+.8);
\draw[->] (3.25-.9,2.5) -| node[above,pos=0.6] {$6$x$7$x$14$} (4-1.3,0.5+1.2) -- (4.75-1.3,0.5+1.2);

\draw[->] (5-1.3,1.5) -- node[above,pos=0.6] {$6$x$14$x$14$} (5.5-1.3,1.5);
\draw[fill=white] (5.5+7*0.05-1.3,7*0.05+1) rectangle (6.25+7*0.05-1.3,0.75+7*0.05+1);
\draw (6.25-1.3,1.5+1)node[txt] (conv1){conv2};
\foreach \foreach \z in {2,...,0}  {%
 \draw[fill=white] (5.5+\z*0.05-1.3,0.1+\z*0.05+1) rectangle (6.25+\z*0.05-1.3,0.85+\z*0.05+1);
}

\draw[fill=white] (7+7*0.05-1.3,7*0.05+1) rectangle (7.75+7*0.05-1.3,0.75+7*0.05+1);
\draw[->] (6.25-1.3,0.5+1) -- node[above,pos=0.67] {$16$x$10$x$10$} (7-1.3,0.5+1);
\foreach \foreach \z in {2,...,0}  {%
 \draw[fill=white] (7+\z*0.05-1.3,0.1+\z*0.05+1) rectangle (7.75+\z*0.05-1.3,0.75+\z*0.05+1);
}
\draw (7.75-1.3,1.5+1)node[txt] (conv1){pool2};
\draw[<->] (7-0.2-1.3,0.85+0*0.05+1) -- node[left] {$16$} (7+0.3-1.3,0.85+0.5+1);
\draw[->] (7.75-1.3,0.5+1) -- node[above,pos=0.6] {$16$x$5$x$5$} (9-1.5,0.5+1);

\draw[fill=white] (9-1.5,0+1) rectangle (9.3-1.5,1+1);
\draw[->] (9.3-1.5,0.5+1) -- node[above] {$400$} (9.8-1.5,0.5+1);
\draw (9-1.5,1.5+1)node[txt] (conv1){flat};

\draw[fill=white] (9.8-1.5,0+1) rectangle (10.1-1.5,1+1);
\draw[fill=white] (10.6-1.5,0+1) rectangle (10.9-1.5,1+1);
\draw[fill=white] (11.4-1.5,0+1) rectangle (11.7-1.5,1+1);
\draw[->] (10.1-1.5,0.5+1) --node[above] {$120$}(10.6-1.5,0.5+1);
\draw[->] (10.9-1.5,0.5+1) --node[above] {$84$} (11.4-1.5,0.5+1);
\draw (10.5-1.5,1.5+1)node[txt] (conv1){dense1 \hspace{.1cm} 2 \hspace{.5cm} 3};

\draw[->] (11.7-1.5,0.5+1) -- node[above] {$10$} (12-1.5,0.5+1);

\coordinate (item1A) at (-3 ,.5);
\coordinate (item1B) at (-1.2,3);       
\draw [fill=gray!40, opacity=.15](item1A) rectangle (item1B) 
                    node[xshift=-1.5cm, yshift=-2.2cm, opacity=1] {item1};

\coordinate (item1A) at (-1.1 , -.5);
\coordinate (item1B) at (4.25-1.3,1.6);       
\draw [fill=gray!40, opacity=.15](item1A) rectangle (item1B) 
                    node[xshift=-3.5ex, yshift=-1.8cm, opacity=1] {item1};

\coordinate (item2A) at (-1.1, 1.9);
\coordinate (item2B) at (4.25-1.3,3.5); 
 \draw [fill=gray!200, opacity=.15](item2A) rectangle (item2B) 
                    node[xshift=-3.5ex, yshift=-1.5ex, opacity=1] {item2};

\coordinate (item1A) at (4.5-1.3 , 0.5);
\coordinate (item1B) at (10.4,3);       
\draw [fill=gray!40, opacity=.15](item1A) rectangle (item1B) 
                    node[xshift=-3.5ex, yshift=-2.15cm, opacity=1] {item1};

\end{tikzpicture}

%% file: lenet-pipeline.tex
\tikzstyle{txt} = [text centered, inner sep=0pt]

\begin{tikzpicture}[scale=1.0,
z={({0.3cm*cos(45)},{0.3cm*sin(45)})},
>=latex, 
font=\footnotesize,
]


\draw[fill=white] (0.25+7*0.05-.6,7*0.05) rectangle (1+7*0.05-.6,.75+7*0.05);
\foreach \foreach \z in {2,...,0}  {%
 \draw[fill=white] (0.25+\z*0.05-.6,.1+\z*0.05) rectangle (1+\z*0.05-.6,.85+\z*0.05);
}
\draw (0.7-.6,1.5)node[txt] (conv1){conv1};
\draw[->] (-1,0.5) -- node[above] {$1$x$32$x$32$} (0.25-.6,0.5);
\draw[<->] (.25-0.2-.6,.85+0*0.05) -- node[left] {$6$} (.25+0.3-.6,.85+0.5);

\draw[fill=white] (2.5+5*0.05-1.1,5*0.05) rectangle (3.25+5*0.05-1.1,0.85+5*0.05);
\foreach \foreach \z in {2,...,0}  {%
 \draw[fill=white] (2.5+\z*0.05-1.1,0.1+\z*0.05) rectangle (3.25+\z*0.05-1.1,.85+\z*0.05);
}
\draw (3.2-1.1,1.5)node[txt] (conv1){pool1};
\draw[->] (1-.6,0.5) --(1.75-1.1,0.5) |- node[above,pos=0.6] {$6$x$28$x$28$} (2.51-1.1,0.5);
\draw[<->] (2.5-0.2-1.1,0.85+0*0.05) -- node[left] {$6$} (2.5+0.3-1.1,0.85+0.5);

\draw[fill=white] (4.75+7*0.05-1.5,7*0.05) rectangle (5.5+7*0.05-1.5,0.75+7*0.05);
\draw (5.5-1.5,1.5)node[txt] (conv1){conv2};
\draw[->] (3.25-1.1,0.5) -- (4-1.1,0.5) |- node[above,pos=0.6] {$6$x$14$x$14$} (4.75-1.5,0.5);
\foreach \foreach \z in {2,...,0}  {%
 \draw[fill=white] (4.75+\z*0.05-1.5,0.1+\z*0.05) rectangle (5.5+\z*0.05-1.5,0.85+\z*0.05);
}
\draw[<->] (4.75-0.2-1.5,0.85+0*0.05) -- node[left] {$16$} (4.75+0.3-1.5,0.85+0.5);

\draw[fill=white] (7+7*0.05-1.9,7*0.05) rectangle (7.75+7*0.05-1.9,0.75+7*0.05);
\foreach \foreach \z in {2,...,0}  {%
 \draw[fill=white] (7+\z*0.05-1.9,0.1+\z*0.05) rectangle (7.75+\z*0.05-1.9,0.75+\z*0.05);
}
\draw (7.75-1.9,1.5)node[txt] (conv1){pool2};
\draw[<->] (7-0.2-1.9,0.85+0*0.05) -- node[left] {$16$} (7+0.3-1.9,0.85+0.5);
\draw[->] (5.5-1.5,0.5) -- (6.25-1.5,0.5) |- node[above,pos=0.67] {$16$x$10$x$10$} (7-1.9,0.5);

\draw[->] (7.75-1.9,0.5) -- node[above,pos=0.6] {$16$x$5$x$5$} (9-2.3,0.5);

\draw[fill=white] (9-2.3,0) rectangle (9.3-2.3,1);
\draw[->] (9.3-2.3,0.5) -- node[above] {$400$} (9.8-2.3,0.5);
\draw (9-2.3,1.5)node[txt] (conv1){flat};

\draw[fill=white] (9.8-2.3,0) rectangle (10.1-2.3,1);
\draw[fill=white] (10.6-2.3,0) rectangle (10.9-2.3,1);
\draw[fill=white] (11.4-2.3,0) rectangle (11.7-2.3,1);
\draw[->] (10.1-2.3,0.5) --node[above] {$120$}(10.6-2.3,0.5);
\draw[->] (10.9-2.3,0.5) --node[above] {$84$} (11.4-2.3,0.5);
\draw (10.5-2.2,1.5)node[txt] (conv1){dense1 \hspace{.4cm} 2 \hspace{.4cm} 3};

\draw[->] (11.7-2.3,0.5) -- node[above] {$10$} (12-2.3,0.5);

\coordinate (item1A) at (-1 , -.5);
\coordinate (item1B) at (2.8, 2);       
\draw [fill=gray!10, opacity=.15](item1A) rectangle (item1B) 
                    node[xshift=-3.2cm, yshift=-2.3cm, opacity=1] {\textbf{item1}};

\coordinate (item2A) at (3 , -.5); 
\coordinate (item2B) at (6.3 , 2); 
 \draw [fill=gray!100, opacity=.15](item2A) rectangle (item2B) 
                    node[xshift=-2.8cm, yshift=-2.3cm, opacity=1] {\textbf{item2}};

\coordinate (item3A) at (6.5 , -.5);
\coordinate (item3B) at (9.6 , 2); 
 \draw [fill=gray!200, opacity=.15](item3A) rectangle (item3B) 
                    node[xshift=-2.6cm, yshift=-2.3cm, opacity=1] {\textbf{item3}};
\end{tikzpicture}

%% file: extsem.tex
\section{\nnef extension for multiple items}
\label{sec-nnef-ext}
The purpose of this section is to propose a manner to separate the specification of each item  so that any execution of the items respecting the specification  properly encode the global ML model.
To that extent, we propose first to extend the syntax
of \nnef to allow explicit parallelization and then to express the associated semantics with colored \petri nets.

\subsection{Extension for item splitting}
We first need to specify the item on which the description will be implemented. To that end,
we enrich the graph definition with the keyword \emph{graphitem}
to provide the logical id of the HW or SW item. 

\begin{mysyntax}[GraphItem]
  \begin{verbatim}
<graph-definition> ::= <graph-declaration> 
  <graph-declaration-item> <body>

<graph-declaration-item> ::= "graphitem" 
  <identifier> <identifier>"("<identifier-list>")" 
  "-$>$" "("<identifier-list>")"
\end{verbatim}
\end{mysyntax}

\begin{semantic}
  The first \emph{$<$identifier$>$} refers to the item id,
  the second \emph{$<$identifier$>$} is the name of the local node
  and
  elements of the \emph{$<$identifier-list$>$} will be input or output of the graphitem. 
The semantic of the \emph{$<$graph-declaration-item$>$}
is such that all instructions within the 
\emph{$<$body$>$} are executed by the item.
\end{semantic}

We also need to \emph{exchange data} between several items and ensure that those exchanged data are available before computation.
To that end, we introduce a new type of variable, namely \emph{variablesync}.
This references a variable that could be read from or write to another \emph{graphitem}. We use a \emph{fragment} to define this new type.

\begin{mysyntax}[VariableSync]
\begin{verbatim}
fragment variablesync<? = scalar>
    (shape: integer[])  -> ( output: tensor<?> )
\end{verbatim}
\end{mysyntax}

\begin{semantic}
  Each \emph{variablesync} is a shared variable with a unique writer and possibly multiple readers.
  Writer is in charge to transmit the variable via the instruction \emph{send\_var}
  and each reader can access this data via \emph{get\_var} instruction.
\end{semantic}
    
We then define new \nnef instructions to send or get data between several \emph{graphitem}. 

\begin{mysyntax}[get_var]
\begin{verbatim}
fragment get_var<? = scalar>(source : graphitem,
    data  : variablesync) -> ( output: tensor<?> )
\end{verbatim}
\end{mysyntax}

\begin{semantic} \emph{get_var} appears in each reader description.
  The output of this instruction is a local
  variable that gets the content of the shared variable
  and which is available for the caller item.
\emph{Source} is the item that writes and provides the shared variable the name of which is given by
\emph{data}.
\end{semantic}

Similarly, the writer must define the instruction to send a shared variable to other items.
\begin{mysyntax}[send_var]
\begin{verbatim}
fragment send_var<? = scalar>
    (  dest  : graphitem[], data  : scalar) 
    -> ( output: variablesync )
\end{verbatim}
\end{mysyntax}

\begin{semantic} \emph{send\_var} appears in the writer description only.
  It takes as input the list of reader items and the name of the tensor that shall be sent.
  The output tensor is a global \emph{variablesync}
  that will support the \emph{synchronization}.
\end{semantic}

The rest of the \nnef syntax remains unchanged.

\subsection{Splitting an \nnef description into multi-item descriptions}
\label{subsec-split-nnef}
Initially, the DNN is described in a unique \nnef
description as shown in section \ref{sec-nnef}.
Such a description contains 3 types of instructions:
\begin{itemize}
\item definition of input tensor(s);
\item definition of fixed parameters;
\item variables computed by each layer.
\end{itemize}
A splitting consists in partitioning the last type of instructions among the items,
adding the adequate definition(s) of tensors / fixed parameters
and adding the adequate \emph{send_var} / \emph{get_var}.
The composition of item descriptions  shall respect the semantics
of the full \nnef description.

\begin{example}
  \label{ex-multiple-nnef}
  \it
Let us consider the DNN of listing \ref{completeMLMD} with its associated \petri net (figure \ref{fig-net-complete}).
Let us assume that the DNN is allocated on 3 items such that $o_1$, $o_6$, $o_7$ and $out$ are computed on item 1;
  $o_2$, $o_3$ are computed on item 2 and  $o_4$, $o_5$ are computed on item 3.
Thus, the description on the items is given in Listing \ref{MLMID1}.

\begin{figure}[hbt]
   \centering
   \resizebox{.95\linewidth}{!}{ 
\begin{lstlisting}[caption=Complete DNN \nnef,label=completeMLMD]
graph DNN(e1) -> (out)
{
   e1 = external..
   v1 = variable... 'variable1'); v2 = ...; v3 = ...; v4 = ...;
   v5 = ...; v6 = ...; v7 = ...; v8 = ...; v9 = ...; v10 = ...;
   o1 = conv(e1,v1,v2,... ;
   o2 = conv(o1,v3,v4,...; o3 = conv(o2,v5,v6,...;
   o4 = conv(o1,v7,v8,...; o5 = conv(o4,v5,v6,...;
   o6 = concat (o3, o5); o7 = flatten(o6,...;
   out = gemm(o7, v9, v10...; 
}
\end{lstlisting}}
\end{figure}

\begin{figure}[hbt]
      \centering
      \resizebox{.80\linewidth}{!}{\input{petri-DNN}}
      \caption{\petri associated to DNN of listing \ref{completeMLMD}. Initial marking\label{fig-net-complete}}
\end{figure}
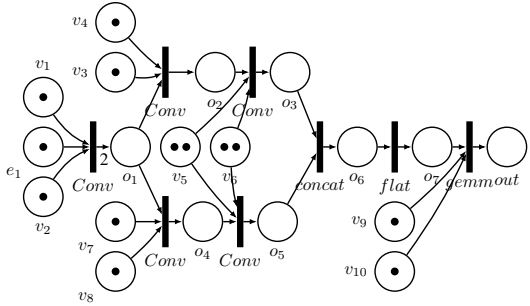


The union of the instructions of each item corresponds to the complete \nnef description with the additional \emph{variablesync}
and the communication instructions.
Locally in the item, the pointers to the input tensor and fixed parameters must also be declared.
\end{example}

\begin{figure}[hbt]
  \centering
   \resizebox{.95\linewidth}{!}{
\begin{lstlisting}[caption=\nnef for all items,label=MLMID1]
graphitem ITEM1 DNN1([e1, vsync2, vsync3]) -> ([vsync1, out])
{
   e1 = external...; 
   v1 = variable... 'variable1'); v2 = ...; v9 = ...;  v10 = ...; 
   vsync1 = variablesync<scalar> (shape ...;
   vsync2 = variablesync ...;   vsync3 = variablesync ...;
    
   o1 = conv(e1,v1,v2,...;
   vsync1 = send_var ([ITEM2, ITEM3], o1);
   o3 = get_var(ITEM2, vsync2);
   o4 = get_var(ITEM3, vsync3);   
   o6 = concat (o3, o4); o7 = flatten(o4,...;
   out = gemm(o7, v9, v10...; 
}

graphitem ITEM2 DNN2([vsync1]) -> ([vsync2])
{
   v3 = variable... 'variable3'); v4 = ...; v5 = ...;  v6 = ...; 
   vsync1 = variablesync ...;  vsync2 = variablesync ...;
    
   o1 = get_var(ITEM1, vsync1)
   o2 = conv(e1,v3,v4,...; o3 = conv(o2,v5,v6,...
   vsync2 = send_var ([ITEM1], o3);
}

graphitem ITEM3 DNN3([vsync1]) -> ([vsync3])
{
   v5 = ...; v6 = ...; v7 = ...; v8 = ...
   vsync1 = variablesync...;  vsync3 = variablesync...;

   o1 = get_var(ITEM1, vsync1)
   o4 = conv(o1,v7,v8,...; o5 = conv(o1,v5,v6,...
   vsync3 = send_var ([ITEM1], o5);
}
\end{lstlisting}}
 \end{figure}

\subsection{\petri-based semantic}
We define the execution model semantics of  multi-item descriptions
using coloured \petri nets \cite{colorpetri}.
We associate a colour to each item where the colour is set to the tokens
and edges (on which the coloured tokens transit).

\begin{translation}
  Let assume there are $N$ items.
  We first apply the translation \ref{trans-nnef-petri} for each item leading to $N$
  independent \petri nets $(P_i,T_i,V_i)$, each with a unique and distinct colour.
For the new instruction, the translation is extended as follows:  
\begin{itemize}
\item a \emph{varsync} does not generate any place;
\item a \emph{get\_var} does not generate any transition;
\item a \emph{send\_var} produces a transition \emph{sync}
  with an incoming edge from the variable the content of which is transmitted.
\end{itemize}

The set of \nnef item descriptions generates a \petri net $(P,T,V)$
which is roughly speaking the union of the $N$ \petri nets $(P_i,T_i,V_i)$.
More precisely,
\begin{itemize}
\item any input tensor or fixed parameter that is duplicated in the \nnef files appears in the \petri net of the item.
  Those duplicated places are merged. Because we use the same naming convention,
  $P=\cup P_i$;
\item the initial tokens in the places are also merged leading to places with possibly multiple tokens and multiple colours;  
\item $T=\cup T_i \cup T'$ where $T'$ are the transitions connecting places of one item to other items thanks to the \emph{sync} transition. More precisely,
\begin{itemize}
\item for each writer, there are $k$ edges back from the \emph{sync} label where $k$ is the number of readers. The colour of the each arrow is the one of the reader and the number of tokens sent back corresponds to the number of time the shared variable appears in the reader item instructions;
\item for each reader, there is an edge from the writer place before \emph{sync} to each transition requiring the shared data;
\end{itemize}
  \item As before, when a coloured token is present in a place, it means that the associated variable is available for the item identified by the colour
and can be used by transition.
\end{itemize}
\end{translation}

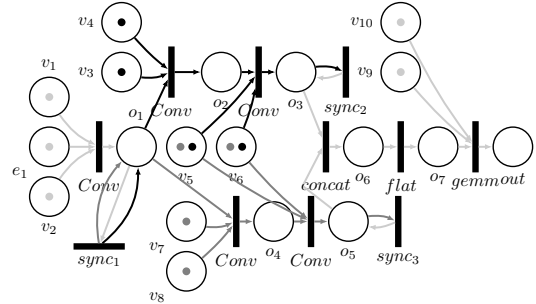
\begin{figure}[hbt]
  \centering
  \resizebox{.8\linewidth}{!}{\input{petri-net-sync}}
  \caption{Semantic of the items synchronization\label{fig:sync}}
\end{figure}

\begin{example}
  \it
The \petri net associated to the example \ref{ex-multiple-nnef} is given in figure \ref{fig:sync}. We present the initial marking with colored tokens. Each color represents the state of one item. Compare to the figure \ref{fig:sem}, we express here the multi items semantics with synchronizations.
\end{example}

\begin{property}[Equivalence between \petri nets]
  The semantics of the multi-items behaviour is equivalent to the complete original ML model.
\end{property}

\begin{remark}
  It is important to explicitly describe the \emph{send_var} and \emph{get_var}
  either in the \nnef files but also in the \petri net
  because items are supposed to be independent
  and segregated.
  In particular, an item X is not allowed to access the memory space of an item Y and interfere with its execution.
  This is classical in aeronautics, see Arinc 653 specification.
  The \xtratum hypervisor \cite{xtratum} is an example of \emph{time and space partitioning} hypervisor that provides communication with \emph{sampling} and \emph{queuing} ports (close to Arinc 653 requirements).
\end{remark}


%% file: petri-DNN.tex
\begin{tikzpicture}[
thick,
node distance=1.1cm,
on grid,
post/.style={-{Latex[length=1.5mm]}},
every transition/.style={fill,minimum width=1mm, minimum height=10mm}
]

\node[place,tokens=1, label=below left:$e_1$] (place1) {};

\node[transition, right=1cm of place1,label=below:$Conv$] (trans0) {};
\node[place,tokens=0,right=.75cm of trans0, label=below:$o_1$] (place0) {};
\node[place,tokens=1,above=1cm of place1, label=above:$v_1$] (placevar10) {};
\node[place,tokens=1,below=1cm of place1, label=below:$v_2$] (placevar30) {};
\draw (placevar10) edge[post, bend right=15] (trans0);
\draw (placevar30) edge[post, bend left=15] (trans0);
\draw (place1) edge[post] (trans0);
\draw (trans0) edge[post] (place0)node[below right] {2};

\node[transition,above right=1.5cm and .7cm of place0,label=below:$Conv$] (trans1) {};
\node[place,tokens=0,right=1cm of trans1, label=below:$o_2$] (place2) {};

\node[place,tokens=1,left=1cm of trans1, label= left:$v_3$] (placevar1) {};
\node[place,tokens=1,above=1cm of placevar1, label= left:$v_4$] (placevar3) {};

\node[transition,right=.75cm of place2,label=below:$Conv$] (trans2) {};
\node[place,tokens=0,right=.75cm of trans2, label=below:$o_3$] (place3) {};
\node[place,tokens=2,below left=1.5cm and .7cm of place2, label=below:$v_5$] (placevar4) {};
\node[place,tokens=2,right=1cm of placevar4, label=below:$v_6$] (placevar5) {};

\draw (place0) edge[post] (trans1);
\draw (trans1) edge[post] (place2);
\draw (place2) edge[post] (trans2);
\draw (trans2) edge[post] (place3);

\draw (placevar1) edge[post, bend right=15] (trans1);
\draw (placevar3) edge[post, bend right=5] (trans1);
\draw (placevar4) edge[post, bend right=5] (trans2);
\draw (placevar5) edge[post, bend right=5] (trans2);

\node[transition,below right=1.5cm and .7cm of place0,label=below:$Conv$] (trans11) {};
\node[place,tokens=0,right=.75cm of trans11, label=below:$o_4$] (place21) {};
\node[place,tokens=1, left=1cm  of trans11, label=below left:$v_7$] (placevar11) {};
\node[place,tokens=1,below=1cm of placevar11, label=below left:$v_8$] (placevar31) {};

\node[transition,right=.75cm of place21,label=below:$Conv$] (trans21) {};
\node[place,tokens=0,right=.75cm of trans21, label=below:$o_5$] (place31) {};

\draw (place0) edge[post] (trans11);
\draw (trans11) edge[post] (place21);
\draw (place21) edge[post] (trans21);
\draw (trans21) edge[post] (place31);
\draw (placevar11) edge[post] (trans11);
\draw (placevar31) edge[post, bend right=5] (trans11);
\draw (placevar4) edge[post, bend right=5] (trans21);
\draw (placevar5) edge[post, bend right=5] (trans21);

\node[transition,below right=1.5cm and .6cm of place3,label=below:$concat$] (trans3) {};
\node[place,tokens=0,right=.75cm of trans3, label=below:$o_6$] (place4) {};

\draw (place3) edge[post] (trans3);
\draw (place31) edge[post] (trans3);
\draw (trans3) edge[post] (place4);

\node[transition, right=.75cm of place4,label=below:$flat$] (trans4) {};
\node[place,tokens=0,right=.75cm of trans4, label=below:$o_7$] (place7) {};
\draw (place4) edge[post] (trans4);
\draw (trans4) edge[post] (place7);

\node[transition, right=.75cm of place7,label=below:$gemm$] (trans5) {};
\node[place,tokens=1, below=1.5cm  of trans4, label=left:$v_9$] (placevar5) {};
\node[place,tokens=1,below=1cm of placevar5, label=left:$v_{10}$] (placevar51) {};
\draw (placevar5) edge[post, bend right=5] (trans5);
\draw (placevar51) edge[post, bend right=5] (trans5);

\node[place,tokens=0,right=.75cm of trans5, label=below:$out$] (place8) {};
\draw (place7) edge[post] (trans5);
\draw (trans5) edge[post] (place8);

\end{tikzpicture}

%% file: petri-net-sync.tex
\begin{tikzpicture}[    
    thick,
    node distance=1.1cm,
    on grid,
    post/.style={-{Latex[length=1.5mm]}},
    tokitem1/.style={colored tokens={gray!40}},
    tokitem2/.style={colored tokens={black}},
    tokitem3/.style={colored tokens={gray!100}},
    tokitem12/.style={colored tokens={gray!100, black}},
    linkitem1/.style={gray!40, line width=0.4mm},
    linkitem2/.style={black, line width=0.4mm},
    linkitem3/.style={gray!100, line width=0.4mm},
    every transition/.style={fill,minimum width=1mm, minimum height=10mm}]

\node[place,tokitem1, label=below left:$e_1$] (place1) {};

\node[transition, right=1cm of place1,label=below:$Conv$] (trans0) {};
\node[place,tokens=0,right=.75cm of trans0, label=above:$o_1$] (place0) {};
\node[place,tokitem1,above=1cm of place1, label=above:$v_1$] (placevar10) {};
\node[place,tokitem1,below=1cm of place1, label=below:$v_2$] (placevar30) {};
\draw[linkitem1]  (placevar10) edge[post, bend right=15] (trans0);
\draw[linkitem1]  (placevar30) edge[post, bend left=15] (trans0);
\draw[linkitem1]  (place1) edge[post] (trans0);
\draw[linkitem1]  (trans0) edge[post] (place0);

\node[transition,below=2cm of trans0,rotate=90,label=left:$sync_1$] (TSYNC) {};
\draw[linkitem3] (TSYNC) edge[post, bend left=25] (place0);
\draw[linkitem2] (TSYNC) edge[post, bend right=25] (place0);
\draw[linkitem1] (place0) edge[post] (TSYNC);

\node[transition,above right=1.5cm and .7cm of place0,label=below:$Conv$] (trans1) {};
\node[place,tokens=0,right=1cm of trans1, label=below:$o_2$] (place2) {};

\node[place,tokitem2, left=1cm of trans1,label=left:$v_3$] (placevar1) {};
\node[place,tokitem2, above=1cm of placevar1, label=left:$v_4$] (placevar3) {};

\node[transition,right=.75cm of place2,label=below:$Conv$] (trans2) {};
\node[place,tokens=0,right=.75cm of trans2, label=below:$o_3$] (place3) {};
\node[place,tokitem12,below left=1.5cm and .7cm of place2, label=below:$v_5$] (placevar4) {};
\node[place,tokitem12,right=1cm of placevar4, label=below:$v_6$] (placevar5) {};

\draw[linkitem2] (place0) edge[post] (trans1);
\draw[linkitem2] (trans1) edge[post] (place2);
\draw[linkitem2] (place2) edge[post] (trans2);
\draw[linkitem2] (trans2) edge[post] (place3);
\draw[linkitem2] (placevar1) edge[post, bend right=15] (trans1);
\draw[linkitem2] (placevar3) edge[post, bend right=5] (trans1);
\draw[linkitem2] (placevar4) edge[post, bend right=5] (trans2);
\draw[linkitem2] (placevar5) edge[post, bend right=5] (trans2);


\node[transition,right=1cm of place3,label=below:$sync_2$] (TSYNC2) {};

\draw[linkitem1] (TSYNC2) edge[post, bend left=15] (place3);
\draw[linkitem2] (place3) edge[post, bend left=15] (TSYNC2);

\node[transition,below right=1.5cm and .6cm of place3,label=below:$concat$] (trans3) {};
\node[place,tokens=0,right=.75cm of trans3, label=below:$o_6$] (place4) {};

\draw[linkitem1] (place3) edge[post] (trans3);
\draw[linkitem1] (trans3) edge[post] (place4);

\node[transition,below right=1.5cm and 2cm of place0,label=below:$Conv$] (trans11) {};
\node[place,tokens=0,right=.75cm of trans11, label=below:$o_4$] (place21) {};
\node[place,tokitem3,left=1cm  of trans11, label=below left:$v_7$] (placevar11) {};
\node[place,tokitem3,below=1cm of placevar11, label=below left:$v_8$] (placevar31) {};

\node[transition,right=.75cm of place21,label=below:$Conv$] (trans21) {};
\node[place,tokens=0,right=.75cm of trans21, label=below:$o_5$] (place31) {};

\draw[linkitem3] (place0) edge[post] (trans11);
\draw[linkitem3] (trans11) edge[post] (place21);
\draw[linkitem3] (place21) edge[post] (trans21);
\draw[linkitem3] (trans21) edge[post] (place31);
\draw[linkitem3] (placevar11) edge[post, bend right=15] (trans11);
\draw[linkitem3] (placevar31) edge[post, bend right=5] (trans11);
\draw[linkitem3] (placevar4) edge[post, bend right=5] (trans21);
\draw[linkitem3] (placevar5) edge[post, bend right=5] (trans21);

\node[transition,right=1cm of place31,label=below:$sync_3$] (TSYNC) {};

\draw[linkitem1] (TSYNC) edge[post, bend left=15] (place31);
\draw[linkitem3] (place31) edge[post, bend left=15] (TSYNC);

\draw[linkitem1] (place31) --($(place31)+(-.9,.7)$);
\draw[linkitem1] ($(place31)+(-.9,.75)$) edge[post] (trans3.west);

\node[transition, right=.75cm of place4,label=below:$flat$] (trans4) {};
\node[place,tokens=0,right=.75cm of trans4, label=below:$o_7$] (place7) {};
\draw[linkitem1]  (place4) edge[post] (trans4);
\draw[linkitem1]  (trans4) edge[post] (place7);

\node[transition, right=.75cm of place7,label=below:$gemm$] (trans5) {};
\node[place,tokitem1, above=1.5cm  of trans4, label=left:$v_9$] (placevar5) {};
\node[place,tokitem1,above=1cm of placevar5, label=left:$v_{10}$] (placevar51) {};
\draw[linkitem1] (placevar5) edge[post, bend right=5] (trans5);
\draw[linkitem1] (placevar51) edge[post, bend right=5] (trans5);

\node[place,tokens=0,right=.75cm of trans5, label=below:$out$] (place8) {};
\draw[linkitem1]  (place7) edge[post] (trans5);
\draw[linkitem1]  (trans5) edge[post] (place8);

\end{tikzpicture}    

%% file: implem.tex
\section{Implementation and experiments}
\label{sec-code-generation}
The previous sections showed how to fulfill the requirements 1 and 2 of the introduction.
The purpose of this section is to give some hint of
how a DO-178C compatible implementation process, as required per requirement 3,
could be defined taking as input an extended \nnef specification.
The considered target, 
a Jetson \xavier TX system-on-chip, 
is composed of
6 Carmel \arm cores, a \gpu, 2 deep learning accelerators (NVDLA) and
other dedicated circuits.
The use of a \nvidia platform is mainly motivated by its availability
and the ease to quickly deploy neural networks application.
We will not discuss
the adequation of \gpu and \cuda implementation with DO-178C objectives because it is an open problem.



The \emph{sync} implementation relies on barrier mechanism
and each item \nnef description is manually coded.
In order to validate the semantic preservation,
we made some instrumentation to show that: 
\begin{enumerate*}[label=(\roman*)]
  \item the execution trace is included in all possible execution traces defined by the \petri net;
  \item the numerical precision is kept;
  \item the measured execution time does not vary.
\end{enumerate*}
We will use the multi-items example presented in example \ref{ex-multiple-nnef} as the specification.
For our experiments, each item is allocated
to one CPU and all the \cuda cores of the \gpu (grouped in a \cuda stream).
As a consequence, we do not guarantee a segregation between items (as they share the GPU)
%
we instead focus on a way to implement
parallel operations of neural networks
with a static code scheduling while preserving the semantics.
As for the \petri semantics, we only developed a code for a single inference.
Nevertheless, it is easy to slightly modify the code by adding loops to handle several input tensors.

\subsection{Get/Send specification}
We chose to implement \emph{get_var} and \emph{send_var}
with
1) global variables stored in the SRAM of the \xavier
and
2) the \posix barrier mechanism of the \emph{pthread} library.
A barrier $b$, shared among several processes,
will block them as long as not all of them reach $b$.
Such a behavior is strictly included in the semantics of the \emph{sync} transition within the \petri net.
However, it is not the most efficient as it prevents
the sending item to proceed until
all the receiving items reach $b$,
whereas
the semantics of \emph{sync} transition only
requires a receiver to wait for the sender (not the sender
 to wait for all receivers).
Nonetheless,
the barrier mechanism is optimal for our  example
because no sender has to process
any instruction before a further \emph{get\_var} or stop execution.

\subsection{Manual code generation}
There are
C and \python interpreters of the \nnef format \cite{nneftools}
but only for traditional CPU target.
Consequently, no existing tool supports our syntax extension nor state-of-the-art \gpu.
Thus we developed the code for each item using C++ and
\cuda using the \cudnn library. Basically the C++ code is executed by the \arm processor whereas  \cuda  allows the definition of kernels that are executed synchronously by all \cuda cores.
The \cudnn library is built on top of \cuda for executing common neural networks layers.

\subsubsection{Software architecture}
Practically,
each type of layer is implemented using a dedicated C++ class that inherits from the abstract \emph{Layer} class
that defines common attributes and methods to be implemented (\emph{init()} and \emph{forward()}) by child classes.
In effect,  \emph{init} statically allocates tensors and \cudnn descriptors while
\emph{forward} launches the layer computation based on \cudnn for Convolution and max pooling layers.
Each item contains one object implementing a static scheduler.
More precisely, during the \emph{init} phase, each item  creates an object for each layer
which are stored in ordered lists.
Thus, items 2 and 3 need one single ordered list whereas item 1 needs two ordered lists (one  for the first part and one for the second part).
During \emph{forward},
the scheduler  calls in order the \emph{forward} of objects stored in ordered lists.

\subsubsection{Scheduling}
We define one separate thread for each item allocated to one CPU + \cuda cores.
More precisely, synchronizations between threads use \texttt{pthread_barrier_t} and associated APIs (\texttt{barrier_init} and \texttt{barrier_wait}). 
Barriers synchronize accesses to shared variables. 

\begin{center}
    \resizebox{.8\linewidth}{!}{\input{sched}}
\end{center}

The execution sequence 
starts with the 3 threads creation on the CPUs
and then reaches the first synchronization barrier.
Then Item1 thread calls the \emph{forward} method of layers of the head (until \emph{send\_var})
while Items 2 and 3 threads wait for the second synchronization barrier.
After,
the second synchronization barrier,
Items 2 and 3 threads  call \emph{forward} method of their layers
while Item1 thread waits for the third synchronization barrier.
After the third synchronization barrier,
Item 1 thread calls \emph{forward} method of layers of the tail.
At last, threads join and exit.

\subsection{Semantic preservation of the \petri net}
\label{sec-petriprese}
The first analysis
aims at verifying that all observed scheduling of layers on the \xavier
respects the \petri net semantics.
Because we use a static scheduler, all schedules should
behave as shown in section \ref{sec-petriprese}
which is included in the semantics of coloured \petri net
of figure \ref{fig:sync}.
For that, we logged each start/end of branches and layers
and we stressed the robustness of the implementation by addind some temporal noises (sleep in the code).

All observed traces respected the schedule of section \ref{sec-petriprese}
with some timing variations.
When observing the implementation with no noise,
execution traces of Item 2 and 3 are interleaved on the \gpu.
When adding a wait of 1s at the beginning of Item 2
(just after barrier1), all layers of Item 3 were executed before those of Item 2. 


\subsection{Semantics preservation of the function}
The second instrumentation mechanism aims at checking
that the functional semantics of the DNN is preserved.
We achieve this by re-implementing the \nnef specification in \pytorch. Then we define 100 random vectors that we run
both on the \pytorch implementation and on the C++ implementation on the \nvidia target. Finally, we compute the overall average error mean between both executions for the 100 runs. 

We were not able to find the exact convolution algorithm of \pytorch. We think that it exists a non documented heuristic that calls the best algorithm (considering execution time) depending of the convolution parameters and available hardware. 
According to the \cudnn documentation \cite{nvidiacudnn},
it is possible to select the convolution algorithm among a list, but details of the implementation are not given.
Thus, there may be a discrepancy between convolutions that we cannot fix.
The average error mean
is extremely small $1.10^{-7}$ for FLOAT32
using 3 \cudnn algorithms (namely gemm, Winograd and direct).
Nevertheless numeric precision results for this experiment are in an acceptable range that is very close to the available numeric precision of floating point representation
and
this also is observed by other frameworks \cite{SilvaCGP22}.

\subsection{Measured Execution Time (MET)}
One objective of the DO-178C that we did not mention until now
is the capacity to estimate the Worst Case Execution Time (WCET). 
Due to the complexity of \nvidia target, a formal demonstration using static analysis 
may be difficult. But at least, a good property is a low variation of the measured execution time among several executions.
In our case, the generated code does not contain any IF-THEN-ELSE patterns
or dynamic loop conditions. Thus, the variability is only linked to the hardware behavior. 
We measured the MET of the complete DNN and of the first convolution of Item 1 over 10 runs.
We rely on the \emph{nsys} tool from \nvidia to get timing measurements.

\begin{center}
  \footnotesize
  \scriptsize

\begin{tabular}{|l|c|c|c|c|}
      \hline
       & \textbf{Mean(MET)} & \textbf{MIN(MET)} & \textbf{(MAX(MET))} & \textbf{STD(MET)} \\
      \hline
      \textbf{First Conv}  & 324 2976 ns  & 322 688 ns & 326 656 ns  & 1.45 ns \\
      \textbf{DNN}  &  24 257 $\mu$s &  16 285 $\mu$s &  53 950 $\mu$s & 13 526 $\mu$s\\
      \hline
  \end{tabular}
\end{center}

The MET of the first convolution is very stable with a very low jitter.
The MET distribution of the DNN is large
and to understand why, we need to investigate the low level behaviour.
\nvidia\ \gpus are  black-boxes processors on which we cannot guarantee worst-case execution time \cite{AmertA21,carle-erts22}.



%% file: sched.tex
\begin{tikzpicture}[xscale=1,transform shape]

    \draw [-latex](-0.5,0) coordinate(dd)-- (0,0) coordinate (O1) -- (12,0)coordinate(ff) node[above]{$t$};
    \draw [dashed,thick] (O1) -- (0,0) coordinate(T1) -- (0,1) coordinate(T2) -- (0,2) coordinate(T3) -- ++(0,.8)coordinate(ff2);
    
    \foreach \nn in{T1,T2,T3}{
        \draw [thick] (dd|-\nn) node[above]{\nn}-- (\nn-|ff);
    }
    
    
    
    \begin{scope}[shift={(T3)}]
        \node[ above right=0.25cm and 0cm of T3,right,draw, minimum width=1.5cm,minimum height=0.5cm,fill=white](T3start) {init};
        \node[ right=0cm of T3start,right,draw, minimum width=0.6cm,minimum height=0.5cm,fill=gray!50](T3w1) {};
        \node[ right=0cm of T3w1,right,draw, minimum width=2.5cm,minimum height=0.5cm,fill=gray!50](T3w2) {};
        \node[ right=0cm of T3w2,right,draw, minimum width=2.5cm,minimum height=0.5cm,fill=white](T3branch) {Item 3};
        \node[ right=0cm of T3branch,right,draw, minimum width=0.5cm,minimum height=0.5cm,fill=gray!50](T3w3) {};
        \node[ right=0cm of T3w3,right,draw, minimum width=3cm,minimum height=0.5cm,fill=gray!50](T3w4) {};
        \draw[-latex,thick] (T3start.north west) -- ++(0,0.2)node[right]{\emph{create}};
    \end{scope}
    
    \begin{scope}[shift={(T2)}]
        \node[ above right=0.25cm and 0.2cm of T2,right,draw, minimum width=1.5cm,minimum height=0.5cm,fill=white](T2start) {init};
        \node[ right=0cm of T2start,right,draw, minimum width=0.4cm,minimum height=0.5cm,fill=gray!50](T2w1) {};
        \node[ right=0cm of T2w1,right,draw, minimum width=2.5cm,minimum height=0.5cm,fill=gray!50](T2w2) {};
        \node[ right=0cm of T2w2,right,draw, minimum width=3cm,minimum height=0.5cm,fill=white](T2branch) {Item 2};
        \node[ right=0cm of T2branch,right,draw, minimum width=3cm,minimum height=0.5cm,fill=gray!50](T2w3) {};
        \draw[-latex,thick] (T2start.north west) -- ++(0,0.2)node[right]{\emph{create}};
    \end{scope}

    \begin{scope}[shift={(T1)}]
        \node[ above right=0.25cm and 0.1cm of T1,right,draw, minimum width=2.0cm,minimum height=0.5cm,fill=white](T1start) {init};
        \draw[-latex,thick] (T1start.north west) -- ++(0,0.2)node[right]{\emph{create}};
        \node[ right=0cm of T1start,right,draw, minimum width=2.5cm,minimum height=0.5cm,fill=white](T1neck) {head};
        \node[ right=0cm of T1neck,right,draw, minimum width=3cm,minimum height=0.5cm,fill=gray!50](T1w1) {};
        \node[ right=0cm of T1w1,right,draw, minimum width=3cm,minimum height=0.5cm,fill=white](T1head) {tail};
    \end{scope}
    
    \draw[line width=1mm] (T1start.south east) -- (T1start.south east |- ff2)node[above]{barrier1};
    \draw[line width=1mm] (T1neck.south east) -- (T1neck.south east |- ff2)node[above]{barrier2};
    \draw[line width=1mm] (T1w1.south east) -- (T1w1.south east |- ff2)node[above]{barrier3};
    \draw[line width=1mm] (T1head.south east) -- (T1head.south east |- ff2)node[above]{join};


\end{tikzpicture}

%% file: related-work.tex
\section{Related work}
\label{sec-related-work}

%
There are plenty of different formats
but no consensus within the community.
State-of-the-art frameworks like \pytorch, or \tensorflow 
rely
on custom black-box formats built on top of \emph{protocol buffers}
\cite{protobuf} developed by Google.
A \emph{protocol buffer} is a structured binary format that is not human readable and
requires conversion tools and template files to be interpreted.
Thus, the syntax is not formally defined and specification of layers are only available through documentation website.
For example, \tensorflow proposes the \emph{.h5} \cite{h5format} format
and  \keras format \cite{kerasformat} builds on top of protocol buffer.
Moreover the way to save a neural network is not unified among frameworks
and may evolve every updated version with poor backward compatibility.
When moving from caffe to caffe2 (known today as \pytorch),  the caffe \cite{caffe} format was not supported anymore.

All previous formats were developed specifically
for training frameworks (open source or proprietary) without any objective for sharing models.
Their main purpose was to allow saving and reloading
previous trained models without too much consideration on syntax and semantics.
\onnx \cite{onnx} and \nnef \cite{nnefformat} were developed with the objective to be independent from frameworks. \onnx is still based on binary protocol buffers \cite{protobuf} (so without a textual syntax) but is proposing a functional semantics through its github site. On the contrary \nnef is proposing a textual format with a syntax and a semantics that is formally defined in a specification.
Unfortunately, the \nnef format suffers from a small community and tools supporting the format.

NNet \cite{nnefformat} format is an example of ad-hoc format.
\reluplex neural networks examples \cite{KatzBDJK17} are in NNet.
It is based on textual files but without definition of syntax and semantics. Import and export tools are provided, but its utilization for sharing neural networks between teams remains supremely painful.

Some other formats like \cite{n2d2,tvm,MLIR}  tackle the need for intermediate representation between a neural network description and an implementation on a target. Especially this supports different optimizations passes like layers folding or low level tensor manipulation description like in LLVM \cite{llvm}. We consider that we are closest to programming language than from a neural network description format. Most of the time \onnx or \nnef are used as input like in \cite{LattnerAB0DPRSV21,pompougnac:hal-03043623,Jin2020}.

Because DNN are data-flow, it is natural to translate them into \emph{synchronous languages}.
There are some works such as
\cite{9094536} that proposed to encode them as Synchronous Dataflow Graphs
or
\scade tool suite \cite{ColacoPPP18} which is currently
developing a DNN libraries.
Once the translation is done, it is then possible
to reuse all the qualified code generation tools.
This is complementary to our work as we could use the \nnef description to generate the \scade program for instance.
To the best of our knowledge, none of actual available neural network description formats propose solution for describing multi items implementation with concerns on sharing variables among them.

%% file: conclusion.tex
\section{Conclusion}
We have proposed a formal extension of \nnef that takes
into account the \emph{execution model} of a description and
allows for the modification of a description of a trained model
to define  traceable distribution
and parallelisation optimizations that preserves the semantics
while improving the execution time compared to a pure sequential aproach.
We have also proposed a code generation strategy based on barriers
for exchanging data between items.
As a future work, a working group has been set up to propose an ONNX aeronautics profile.